\documentclass[letterpaper]{article}
\usepackage[preprint]{liveevalstyle}
\usepackage[hyphens]{url}
\usepackage{natbib}
\usepackage{caption}
\usepackage{microtype}

\usepackage{graphicx}

\usepackage{enumitem}

\usepackage{amsmath} 
\usepackage{amssymb}
\usepackage{booktabs}
\usepackage{multirow}
\usepackage{tikz}
\usepackage{bm}
\usepackage{color}
\usepackage{xcolor}
\usepackage{colortbl}
\usepackage{longtable}
\usepackage{tabularx}
\usepackage{array}

\usepackage{listings}
\usepackage{tcolorbox}
\tcbuselibrary{listings, skins, breakable}

\newtcblisting{promptbox}[1][]{
  colback=gray!5,
  colframe=gray!40,
  arc=6pt,
  boxrule=0.5pt,
  left=8pt,
  right=8pt,
  top=6pt,
  bottom=6pt,
  title=#1,
  coltitle=black,
  fonttitle=\small\bfseries\sffamily,
  attach boxed title to top left={xshift=8pt, yshift=-3pt},
  boxed title style={colback=gray!5, colframe=gray!40, arc=4pt, boxrule=0.5pt},
  listing only,
  listing options={
    basicstyle=\small\ttfamily,
    breaklines=true,
    aboveskip=0pt,
    belowskip=0pt,
    showstringspaces=false,
    columns=fullflexible,
    resetmargins=true,
    xleftmargin=0pt,
    xrightmargin=0pt,
    framexleftmargin=0pt,
    framexrightmargin=0pt,
  }
}

\newcommand{\revdel}[1]{}
\newcommand{\revadd}[1]{#1}
\newcommand{\revblockcolor}{\color{black}}

\title{LiveEvalBench: Toward Open-World Evaluation for Web Generation}

\author{
    Yiyao Wang\textsuperscript{\rm 1}\thanks{Yiyao Wang, Zhen Wen, Yinghao Tang, and Yixiao Fu are with the State Key Lab of CAD\&CG, Zhejiang University. E-mail: \{wangyiyao, wenzhen, yinghaotang, 3210101100\}@zju.edu.cn.},
    Zhen Wen\textsuperscript{\rm 1},
    Yinghao Tang\textsuperscript{\rm 1},
    Yixiao Fu\textsuperscript{\rm 1},
    Lin Yuan\textsuperscript{\rm 2}\thanks{Lin Yuan, Xiaolu Zhang, and Jun Zhou are with Ant Group.},
    Xiaolu Zhang\textsuperscript{\rm 2},
    Jun Zhou\textsuperscript{\rm 2},
    Wei Chen\textsuperscript{\rm 3}\thanks{Wei Chen is with the State Key Lab of CAD\&CG, Zhejiang University, and also with the Laboratory of Art and Archaeology Image (Zhejiang University), Ministry of Education, China. E-mail: chenvis@zju.edu.cn.}
}

\affiliations{
    \textsuperscript{\rm 1}State Key Lab of CAD\&CG, Zhejiang University\\
    \textsuperscript{\rm 2}Ant Group\\
    \textsuperscript{\rm 3}State Key Lab of CAD\&CG, Zhejiang University; Laboratory of Art and Archaeology Image (Zhejiang University), Ministry of Education, China
}

\begin{document}
\maketitle

\begin{abstract}
Large language models are increasingly capable of synthesizing executable frontend projects, yet existing benchmarks still treat web generation as a static evaluation problem. We argue that frontend artifacts demand a different paradigm: they are interactive rather than static, admit diverse yet equally valid implementations, and evolve faster than rigid pipelines can accommodate. To address these gaps, we present LiveEvalBench, an automated framework that reformulates web-generation evaluation as an agentic, adaptive, and extensible process. LiveEvalBench instantiates evaluation as a collaborative review workflow, in which a Build Engineer, a Code Engineer, and a UI Tester collectively gather evidence across the full lifecycle of a frontend project, from deployment and code inspection to browser-based interaction. To handle implementation diversity, an adaptive protocol couples shared rubrics for cross-model comparability with implementation-grounded criteria tailored to each artifact. The framework further supports incremental integration of new evaluator roles and assessment dimensions without pipeline redesign. Experiments across diverse real-world web-generation scenarios show that LiveEvalBench aligns closely with human expert judgment and provides fine-grained insights into frontier models' web generation capabilities. Code is available at \url{https://github.com/wyysteelhead/LiveEvalBench}.
\end{abstract}


\section{Introduction}
\label{sec:intro}

\begin{figure}[!t]
    \centering
    \includegraphics[width=.82\columnwidth]{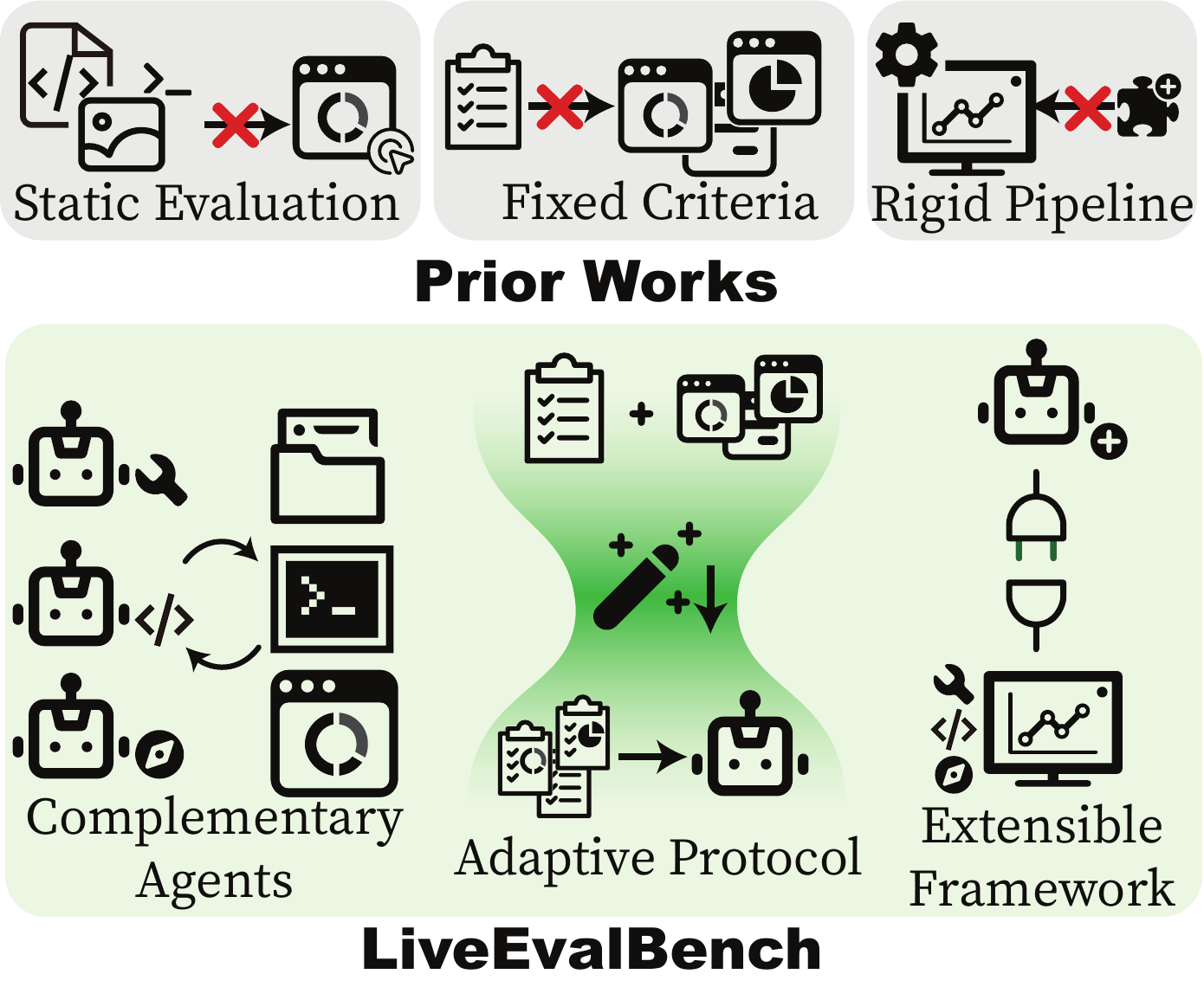}
    \caption{LiveEvalBench versus prior web coding benchmarks. While prior works treat evaluation as a one-shot scoring procedure, LiveEvalBench models it as a collaborative expert review workflow with complementary agents, an adaptive protocol, and an extensible framework.}
    \label{fig:teaser}
\end{figure}

Large language models (LLMs) are rapidly evolving from generating isolated code snippets to synthesizing complete, executable frontend projects \citep{si2025design2code,zhu2025frontendbench,wu2026frontalk,tran2026vibecodebench}. As project-level web generation becomes a practical capability of modern LLMs, a fundamental question emerges: \emph{how should we evaluate what models actually build?}

Despite this progress, evaluation has not kept pace. Existing benchmarks for web generation inherit assumptions from traditional code evaluation: they assess outputs through predefined criteria, fixed execution scripts, or static judgments over source code and screenshots \citep{si2025design2code,zhu2025frontendbench,zhang2025artifactsbench,he2026vision2web}. However, frontend applications are not static outputs. They are interactive, admit diverse yet equally valid implementations, and continuously evolve over time. This creates a fundamental paradigm mismatch: (1) static evaluation cannot judge artifacts whose quality depends on runtime interaction; (2) fixed rubrics cannot fairly assess open-ended generation where the same query admits many valid solutions; and (3) rigid pipelines cannot keep up with the rapid evolution of web generation capabilities.

In this paper, we present \textbf{LiveEvalBench}, an automated evaluation framework designed for open-world web generation. Rather than treating evaluation as a one-shot scoring procedure, LiveEvalBench models it as a collaborative expert review workflow that is \emph{agentic}, \emph{adaptive}, and \emph{extensible} (Fig.~\ref{fig:teaser}).

First, to evaluate interactive artifacts through interaction rather than static inspection, we instantiate evaluation as a multi-agent workflow. A \textit{Build Engineer} deploys the project and produces runtime artifacts. A \textit{Code Engineer} inspects implementation quality and instruction following from the source side. A \textit{UI Tester} actively explores the running application through browser-based interaction. Each agent operates with distinct tools and evidence channels, collectively covering the full lifecycle of a frontend project from deployment to code to live behavior.

Second, to fairly evaluate open-ended generation where the same user request admits diverse implementations, we propose an \emph{adaptive evaluation protocol}. Shared rubric items are fixed across all models answering the same query, preserving cross-model comparability. On top of these, implementation-grounded checks are synthesized for each generated project based on what the model actually built, so that scoring reflects the artifact's true capability rather than penalizing valid but unexpected design choices.

Third, to support an evaluation framework that evolves alongside web generation itself, we design an \emph{extensible infrastructure}. Each evaluator is declared as a configuration of three components: a persona specifying who is judging, a set of criteria specifying what is judged, and a set of tools specifying what evidence is consulted. Adding a new evaluation perspective (e.g., a mobile-only user, a color-blind user) requires filling in this configuration rather than redesigning the pipeline.

We instantiate LiveEvalBench on a benchmark of \revdel{$45$}\revadd{$100$} real-world web generation queries evaluated across $11$ frontier models. \revdel{The resulting evaluation yields concrete findings about frontier models: interactive runtime behavior consistently emerges as the dominant source of failure across all models, and different models exhibit markedly different per-dimension failure profiles.} \revadd{The resulting evaluation shows that current frontier models fail most often on interactive runtime behavior.} \revdel{Our framework reaches substantial agreement} \revdel{with human experts ($86.8\%$, Cohen's $\kappa = 0.60$).} \revadd{Across multiple validation experiments, LiveEvalBench shows strong alignment with human judgment across evaluation dimensions.}

Our contributions are summarized as follows:

\begin{enumerate}
    \item We identify a fundamental evaluation paradigm mismatch in web generation and propose \textbf{LiveEvalBench}, an \textbf{agentic evaluation framework} for multi-perspective assessment using build, code, and browser evidence.

    \item We introduce an \textbf{adaptive evaluation protocol} that balances cross-model comparability with implementation-aware scoring for open-ended web generation.

    \item We develop an \textbf{extensible evaluation infrastructure} where new evaluator roles and assessment dimensions can be added through configuration without redesign.

    \item We construct a benchmark for web generation and uncover strengths and weaknesses of frontier models overlooked by existing evaluations.

\end{enumerate}

\section{Related Work}
\label{sec:related}

\begin{table*}[t]
\centering
\setlength{\tabcolsep}{3.2pt}
\renewcommand{\arraystretch}{1.1}
\newcommand{\cmark}{\checkmark}
\newcommand{\pmark}{$\triangle$}
\newcommand{\nmark}{\textemdash}
\begin{tabular}{lccccccccc}
\toprule
\textbf{Work} & \multicolumn{5}{c}{\textbf{Perspectives}} & \multicolumn{2}{c}{\textbf{Method}} & \multicolumn{2}{c}{\textbf{Design}} \\
\cmidrule(lr){2-6}\cmidrule(lr){7-8}\cmidrule(lr){9-10}
& \textbf{Build} & \textbf{Src.} & \textbf{Brwsr.} & \textbf{Interact.} & \textbf{Visual} & \textbf{Exec.} & \textbf{Adapt.} & \textbf{Multi-Eval.} & \textbf{Extens.} \\
\midrule
HumanEval \citep{chen2021evaluating} & \nmark & \cmark & \nmark & \nmark & \nmark & \cmark & \nmark & \nmark & \nmark \\
SWE-bench \citep{jimenez2024swebench} & \pmark & \cmark & \nmark & \nmark & \nmark & \cmark & \nmark & \nmark & \nmark \\
Design2Code \citep{si2025design2code} & \nmark & \nmark & \pmark & \nmark & \cmark & \nmark & \nmark & \nmark & \nmark \\
Web-Bench \citep{xu2025webbench} & \pmark & \cmark & \cmark & \cmark & \nmark & \cmark & \nmark & \nmark & \nmark \\
FrontendBench \citep{zhu2025frontendbench} & \nmark & \cmark & \cmark & \cmark & \pmark & \cmark & \nmark & \nmark & \nmark \\
ArtifactsBench \citep{zhang2025artifactsbench} & \nmark & \pmark & \cmark & \cmark & \cmark & \pmark & \pmark & \nmark & \nmark \\
FronTalk \citep{wu2026frontalk} & \nmark & \nmark & \cmark & \cmark & \cmark & \cmark & \pmark & \pmark & \nmark \\
Vision2Web \citep{he2026vision2web} & \pmark & \nmark & \cmark & \cmark & \cmark & \cmark & \pmark & \cmark & \nmark \\
Vibe Code Bench \citep{tran2026vibecodebench} & \pmark & \nmark & \cmark & \cmark & \pmark & \cmark & \nmark & \cmark & \nmark \\
WebCompass \citep{lei2026webcompass} & \pmark & \pmark & \cmark & \cmark & \cmark & \cmark & \cmark & \cmark & \nmark \\
\textbf{LiveEvalBench} & \cmark & \cmark & \cmark & \cmark & \cmark & \cmark & \cmark & \cmark & \cmark \\
\bottomrule
\end{tabular}
\caption{Comparison of representative code, web, and frontend evaluation benchmarks. \cmark indicates explicit support, \pmark indicates partial or indirect support, and \nmark indicates that the capability is not a primary part of the evaluation protocol. Src., Brwsr., Interact., Adapt., Multi-Eval., and Extens. denote source, browser, interaction, adaptive, multi-evaluator, and extensible, respectively.\revdel{ The columns reflect the three contributions of LiveEvalBench: an \emph{agentic evaluation framework} that collects multi-perspective evidence through multiple evaluator agents, an \emph{adaptive evaluation protocol} that pairs shared rubric items with implementation-grounded checks, and an \emph{extensible evaluation infrastructure} that allows new evaluator roles and assessment dimensions to be added without rewriting the pipeline.}}
\label{tab:related_work_comparison}
\end{table*}

\subsection{Benchmark for Web Generation}

We organize prior work by the kind of evidence the evaluator inspects: source code, runtime functionality, and rendered visuals.

\paragraph{Code evaluation}
Pure code-generation benchmarks score programs by executing them against fixed unit tests, spanning function-level suites \citep{chen2021evaluating,austin2021mbpp,hendrycks2021apps,liu2023evalplus}, contamination-resistant and library-rich variants \citep{jain2025livecodebench,zhuo2025bigcodebench}, and repository-level engineering settings \citep{jimenez2024swebench}. Web-generation benchmarks that adopt code-level checks face two complications: the artifact is a frontend project rather than a single program, so the ``code'' channel must also cover build configuration and asset organization; and ground-truth tests written against a fixed reference no longer apply when many UI layouts can satisfy the same user request \citep{xu2025webbench,lei2026webcompass}.\revdel{ LiveEvalBench therefore treats source-code evidence as one channel among several.}

\paragraph{Functional evaluation}
A second line evaluates whether the deployed webpage behaves correctly at runtime. Sandboxed automatic-test pipelines pair each task with predefined scripts executed in a controlled environment \citep{xu2025webbench,zhu2025frontendbench}, achieving reproducibility at the cost of binding tests, fixtures, and judging logic to specific task templates. A more recent category lets an evaluator agent explore the deployed application in a real browser instead of executing fixed scripts \citep{wu2026frontalk,he2026vision2web,lu2025webgenbench,tran2026vibecodebench}. WebCompass \citep{lei2026webcompass} is representative, using an agent-as-a-judge procedure that synthesizes targeted test cases in the browser during execution. Adjacent web-agent resources \citep{deng2023mind2web,zhou2024webarena,koh2024visualwebarena} do not evaluate generated frontends but inform evaluator design.

\paragraph{Visual evaluation}
A third line judges the rendered interface. One branch treats UI generation as visual translation and scores generations against a target design \citep{beltramelli2018pix2code,yun2024web2code,li2025sketch2code}: Design2Code \citep{si2025design2code} is representative, benchmarking multimodal models on reproducing real-world webpages from screenshots via visual similarity and human preference. A second branch renders the artifact and asks an MLLM judge to score it via a per-task checklist \citep{wan2024interaction2code}: ArtifactsBench \citep{zhang2025artifactsbench} is representative, driving each artifact through scripted interactions and scoring visual and interactive integrity. Both branches use rubrics tied to a single target or task template.

\paragraph{\revadd{Limitations of Existing Benchmarks}}
Across these efforts, existing benchmarks have made important progress in web generation evaluation, but two limitations remain. First, their adaptive capacity is often bounded by predefined evaluation artifacts (e.g., expert-defined checklists), which ties evaluation to the benchmark's existing queries and makes it costly to evaluate newly emerging web generation scenarios. Second, their evaluation pipelines are usually rigid: adding a new evaluator role or assessment dimension requires redesigning the pipeline rather than extending the existing framework. LiveEvalBench addresses these limitations within a unified \emph{agentic evaluation framework}, using an \emph{adaptive evaluation protocol} that grounds shared rubric items into implementation-specific checks and an \emph{extensible evaluation infrastructure} that supports new evaluator roles and assessment dimensions without redesigning the pipeline.

\subsection{LLM-as-Judge and Multi-Evaluator Frameworks}

LLM-as-judge work spans single proprietary judges with structured rubrics \citep{zheng2023mtbench,liu2023geval,liu2024alignbench}, specialized open evaluator models \citep{kim2024prometheus,kim2024prometheus2}, studies of judge reliability, bias, and validity \citep{chen2025distributional,guerdan2025indeterminacy,bean2025constructvalidity}, and survey-level framings \citep{li2025generationtojudgment}. Two threads directly motivate our design: \emph{multi-judge juries} replace one large judge with heterogeneous smaller judges, matching or outperforming a single strong judge while reducing bias and cost \citep{verga2024juries,dubois2023alpacafarm}; and \emph{agent-as-a-judge} equips evaluators with tools to inspect files, execute code, or browse pages, approaching human reliability on coding tasks and materially affecting outcomes in browser-based web evaluation \citep{zhuge2024agentasjudge,tran2026vibecodebench}. Existing panels operate over textual outputs rather than multi-channel evidence, and evaluator behaviors are rarely modularized as swappable components---gaps our agentic evaluation framework and extensible evaluation infrastructure address.

\section{Method}
\label{sec:method}

\begin{figure*}[!t]
\centering
\includegraphics[width=0.9\textwidth]{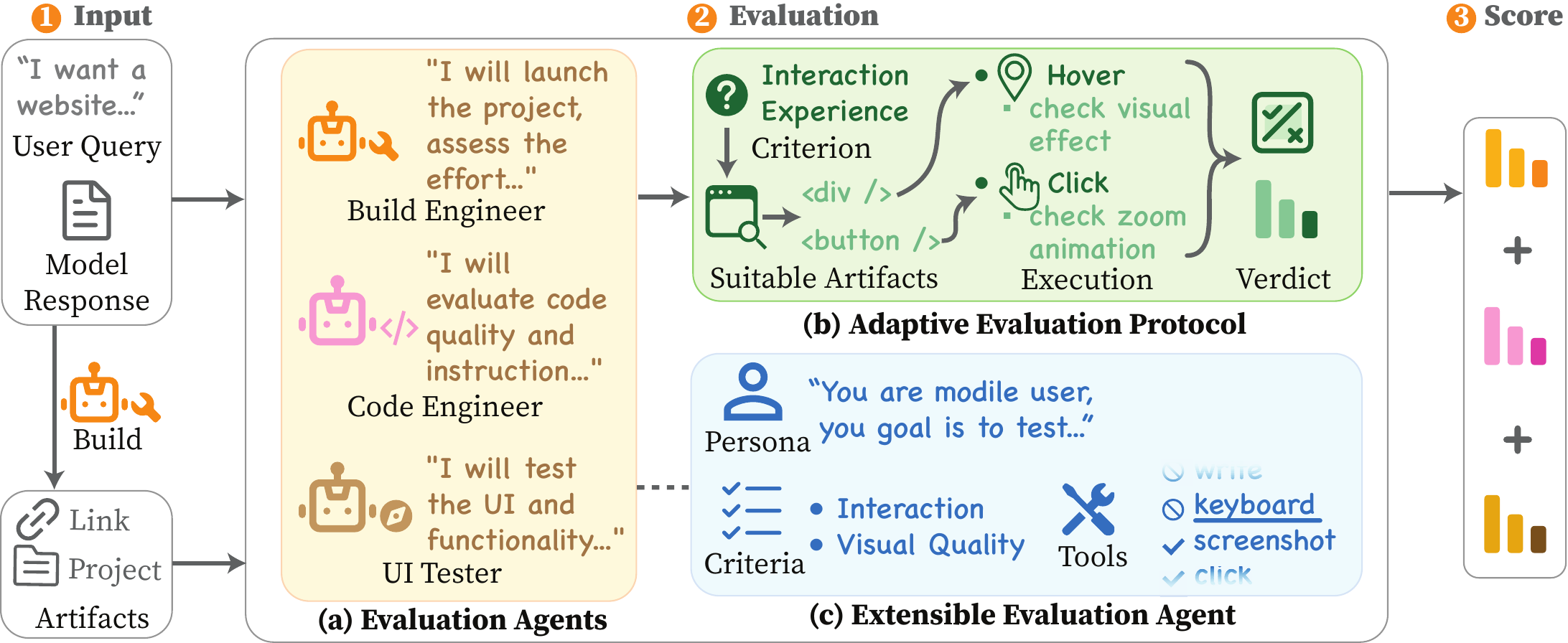}
\caption{Overview of the LiveEvalBench framework. Given a user query and a model response, the response is deployed into a runnable link and a project directory, which are then evaluated in parallel by three specialized agents \textbf{(a)}. Each agent follows an adaptive evaluation protocol that grounds shared rubric criteria into implementation-specific checks at runtime \textbf{(b)}, and the agent design is extensible, supporting new evaluators through configuration \textbf{(c)}.}
\label{fig:framework_overview}
\end{figure*}

\subsection{Framework Overview}

The overview of LiveEvalBench is presented in Fig.~\ref{fig:framework_overview}. LiveEvalBench takes the user query and model response as input, and deploys the response into a runnable link and a deployed project on disk. These artifacts are then handed to three specialized evaluation agents, namely the Build Engineer, the Code Engineer, and the UI Tester, which examine complementary facets of the project in parallel (Fig.~\ref{fig:framework_overview}(a)). During examination, each agent follows an adaptive evaluation protocol that grounds shared rubric criteria into implementation-specific checks at runtime (Fig.~\ref{fig:framework_overview}(b)). The framework is also extensible, supporting new evaluator roles through a unified agent infrastructure (Fig.~\ref{fig:framework_overview}(c)). The resulting per-criterion judgments are aggregated into a final 90-point score: Build contributes 15 points, Code contributes 30 points, and UI contributes 45 points, with per-evaluator caps that prevent any single facet from dominating.

\subsection{Evaluation Agents}
\label{subsec:eval_stage}

As shown in Fig.~\ref{fig:framework_overview}(a), LiveEvalBench employs three specialized evaluation agents that examine these complementary facets in parallel, each operating within the environment natural to its perspective.

\paragraph{\textbf{Build Engineer}} The Build Engineer owns the deployment perspective and operates within the project directory together with a shell. It is also responsible for producing the build artifacts in the input step: it extracts the project files, performs minor repairs to auxiliary configuration when necessary, and launches the application following the response's setup instructions. During evaluation, the Build Engineer revisits its own build trajectory (i.e.), the commands it issued and the auxiliary repairs it made, and from this evidence judges how much extra effort the deployment demanded beyond what the model response itself prescribes.

\paragraph{\textbf{Code Engineer}} The Code Engineer takes the perspective of a developer inspecting the implementation, with access to the source tree and the ability to run commands in the terminal for verification. It evaluates the project on the code side along two aspects: implementation quality and instruction following. Implementation quality considers both code readability and robustness, reflecting whether the code is well organized and whether it handles edge cases gracefully. Instruction following checks whether hard requirements stated in the user query are faithfully realized in the implementation, such as the use of a designated framework (e.g., Vue or React).

\paragraph{\textbf{UI Tester}} The UI Tester stands in for the end user and interacts with the running web link solely through the browser, without access to the underlying code. It assesses the project along three criteria: visual quality, interaction experience, and a set of query-specific criteria generated from the user query. Visual quality looks at whether the interface renders completely and presents itself in a polished and visually coherent manner. Interaction experience examines whether the interface responds clearly and flows intuitively under user actions. The query-specific criteria, tailored to what each query actually asks for, provide a thorough examination of functionality.

\subsection{Adaptive Evaluation Protocol}
\label{subsec:adaptive}

As sketched in Fig.~\ref{fig:framework_overview}(b), we decouple \emph{what} each evaluator should check from \emph{how} those checks work. The first concern adapts to the user query; the second adapts to what each model actually produced.

\paragraph{What to check (per query)} Each evaluator carries a set of fixed criteria along the dimensions described above. On top of these, the UI Tester additionally takes in \emph{query-specific criteria}. To faithfully cover every requirement raised by the user, we extract the functional requirements expressed in the query and turn each of them into a corresponding criterion. These query-specific criteria are shared across all models answering the same query, ensuring comparability between models. Whether to take in query-specific criteria is itself a configuration choice; the Code Engineer opts out and relies solely on its fixed rubric.

\paragraph{How to check (per implementation)} A criterion such as \emph{interaction experience} is too abstract to apply directly; it must be turned into concrete actions on this particular project. For each criterion, the evaluator first inspects the implementation through what it can observe. For example, the UI Tester can examine the rendered DOM and accessibility tree of the running project. For each applicable element it finds there, the evaluator then writes an individual executable check tailored to that element. As illustrated in Fig.~\ref{fig:framework_overview}(b), the criterion of interaction experience is grounded in the buttons and containers actually present in the project, and is then verified through hover and click actions tailored to those elements. The criteria themselves stay fixed across all models, while the concrete checks adapt to each implementation.



\subsection{Extensible Evaluation Infrastructure}
\label{subsec:extensible}

For the framework to evolve alongside web generation, adding a new evaluation perspective must be cheap. This reduces to a more basic question: what is the minimum needed to define an evaluator? We summarize it into three components, illustrated in Fig.~\ref{fig:framework_overview}(c) with a touch-only mobile user as a running example. The first is a \textbf{persona} that specifies \emph{who} is judging, fixing the role the evaluator plays and the standards it brings to the interface. The second is a set of \textbf{criteria} that specifies \emph{what} is judged; the mobile user in Fig.~\ref{fig:framework_overview}(c) still cares about interaction and visual quality, the same dimensions the UI tester would examine. What separates the two is the third component: a set of \textbf{tools} that specifies \emph{what evidence} is consulted. The mobile user is barred from using the keyboard and can only interact with the interface by touch; source-level access such as read and write is also denied. Once the triple is filled in, the orchestrator schedules the evaluator and aggregates its score under the same rules as the default jury, with no changes to framework code.

This abstraction extends easily beyond the running example. A color-blind end-user, for instance, can be added by reusing the same persona and criteria slots while swapping in a new screenshot tool that applies a color-vision transform, so the evaluator sees the page as the user would. Other perspectives, such as a visually impaired user relying on the accessibility tree, follow the same recipe by filling in the three slots. The full schema for specifying an evaluator under this three-slot abstraction is provided in the supplementary material.

\section{Benchmark Construction}
\label{sec:benchmark}

\subsection{Query Construction}
\label{subsec:query_construction}

\begin{figure}[t]
\centering
\includegraphics[width=0.8\linewidth]{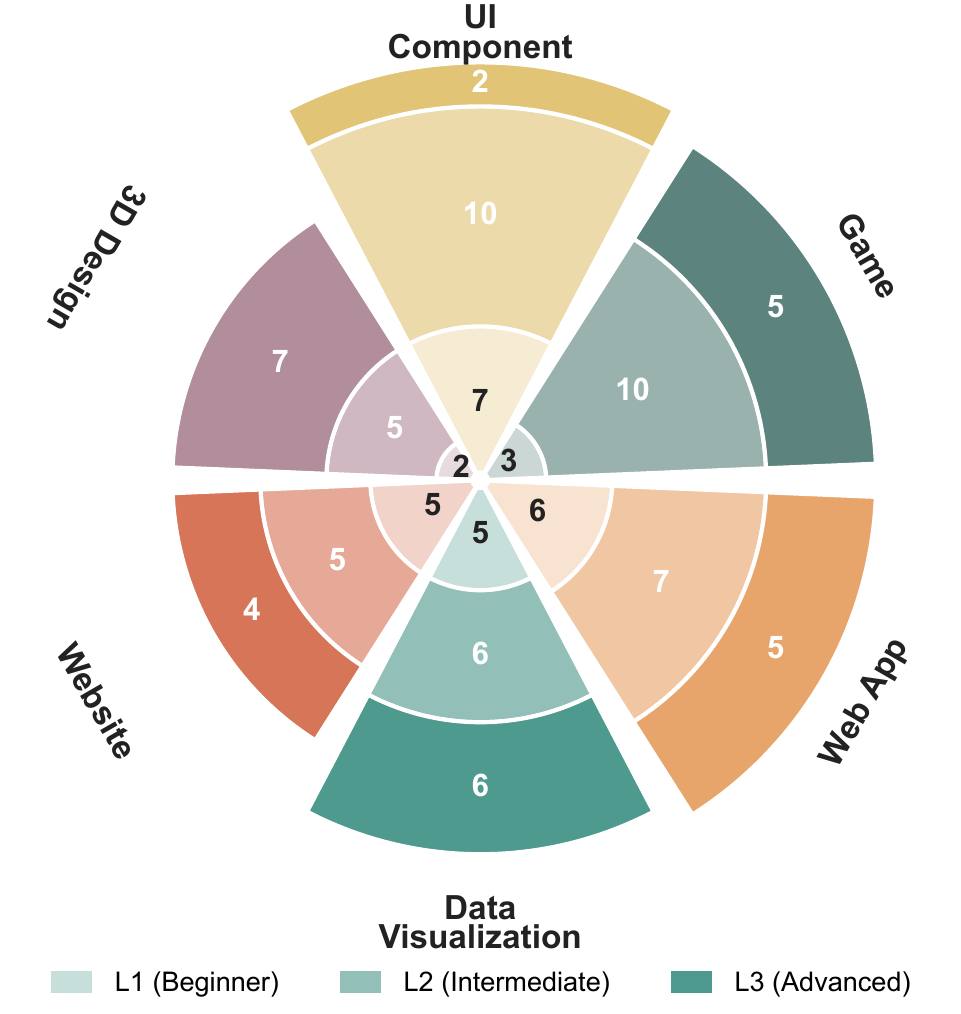}
\caption{Distribution of LiveEvalBench queries across six categories. Each petal is one category, and the three concentric layers break it down by difficulty (L1 inner / L2 middle / L3 outer).}
\label{fig:query_category_dist}
\end{figure}

LiveEvalBench consists of \revdel{45}\revadd{100} frontend project specifications across three difficulty levels (\revdel{20 L1, 20 L2, and 5 L3}\revadd{28 Level-1 (L1), 43 Level-2 (L2), and 29 Level-3 (L3) queries}) and six task categories adapted from ArtifactsBench~\cite{zhang2025artifactsbench} and Design Arena~\cite{designarena2025}: \emph{Data Visualization} (\revdel{10}\revadd{17}), \emph{UI Component} (\revdel{8}\revadd{19}), \emph{Game} (\revdel{8}\revadd{18}), \emph{Web App} (\revdel{8}\revadd{18}), \emph{Website} (\revdel{6}\revadd{14}), and \emph{3D Design} (\revdel{5}\revadd{14}). \revadd{The queries also span three specification granularities: abstract intent-level requests (46), product-requirements-document (PRD)-style specifications (41), and functional descriptions (13).} The joint distribution over categories and difficulty levels is shown in Fig.~\ref{fig:query_category_dist}. We build the benchmark through a four-stage pipeline.

\paragraph{Seed Collection}
We aggregate real-world frontend requests from three complementary sources: public tutorials and coding guides, social media platforms, and task samples from ArtifactsBench~\cite{zhang2025artifactsbench}. Unlike prior benchmarks sourced from sites like GitHub, we treat social media as a primary source, since it is where practitioners share the most up-to-date queries probing frontier code-generation models. We retain only items containing a self-contained, implementable specification, and deduplicate across sources. Seeds serve only as topical anchors and are never used directly.

\paragraph{Role-Conditioned Synthesis}
Real-world frontend specifications differ substantially in abstraction and technical detail. To capture this heterogeneity, we rewrite each seed under four roles with distinct granularity: ``end user'' (intent-driven, non-technical), ``designer'' (visual and interaction details), ``project manager'' (structured requirements and acceptance criteria), and ``developer'' (component decomposition, APIs, edge cases). Rewrites are produced by three frontier models, namely Gemini 3 Pro Preview, GPT-5.2, and Claude Sonnet 4.5, using role-specific prompts provided in the supplementary material. This yields a pool of $\sim$4{,}000 candidate specifications that vary along two orthogonal axes: topical diversity (from seeds) and stylistic diversity (from roles and rewriters).

\paragraph{Difficulty Stratification}
Each candidate is scored by an LLM judge along three rubric dimensions (number of views, external dependency complexity, and interaction/state complexity) and mapped to L1/L2/L3 via a deterministic rule. L1 covers small, self-contained features confined to a single page, such as a single form or an isolated canvas effect; L2 covers moderately complex tasks that may depend on several external libraries, involve a larger codebase, or require non-trivial application logic; L3 covers multi-page or multi-view applications with cross-view navigation, or several intricate features composed together. Two authors verify the labels and resolve disagreements by discussion.

\paragraph{Human Curation}
From the stratified pool we hand-pick the final \revdel{45}\revadd{100} queries to cover all six categories and multiple specification granularities across the benchmark, without enforcing equal counts per cell; selection prioritizes clarity, feasibility, evaluability, and novelty as judged by 2 experts with frontend engineering experience.

The full \revdel{45}\revadd{100} specifications, with granularity, category, difficulty annotations, and per-query evaluation criteria, will be publicly released upon publication under a license permitting free research use.

\subsection{Generated Frontend Projects}
\label{subsec:model_outputs}

\revdel{We run the benchmark queries through $11$ frontier code-capable LLMs, as listed in the Experiments section, to obtain generated frontend projects for evaluation.} \revadd{To instantiate the benchmark with model-generated artifacts, we run each query through the $11$ frontier code-capable LLMs listed in the Experiments section, yielding one generated frontend project per query--model pair.} All models are queried in a single turn with provider-default decoding settings. The only modification to the raw user specification is a single prepended instruction that frames the model as a frontend engineer and asks it to return frontend project content. This minimal-intervention protocol isolates each model's native web generation ability and keeps comparisons faithful to default behavior. \revdel{Environment-level failures and invalid logs are filtered before aggregation; the retained evaluation-record count used for the reported results is given in the Experiments section.}

\subsection{Per-Query Evaluation Criteria}
\label{subsec:per_query_checklists}

LiveEvalBench is released with a set of criteria attached to each query. These criteria are produced following the adaptive evaluation protocol and manually reviewed for faithfulness. The full schema is provided in the supplementary material.

\begin{figure*}[!t] 
  \centering
  
  \begin{minipage}{0.48\textwidth}
    \centering
    \includegraphics[width=0.9\linewidth]{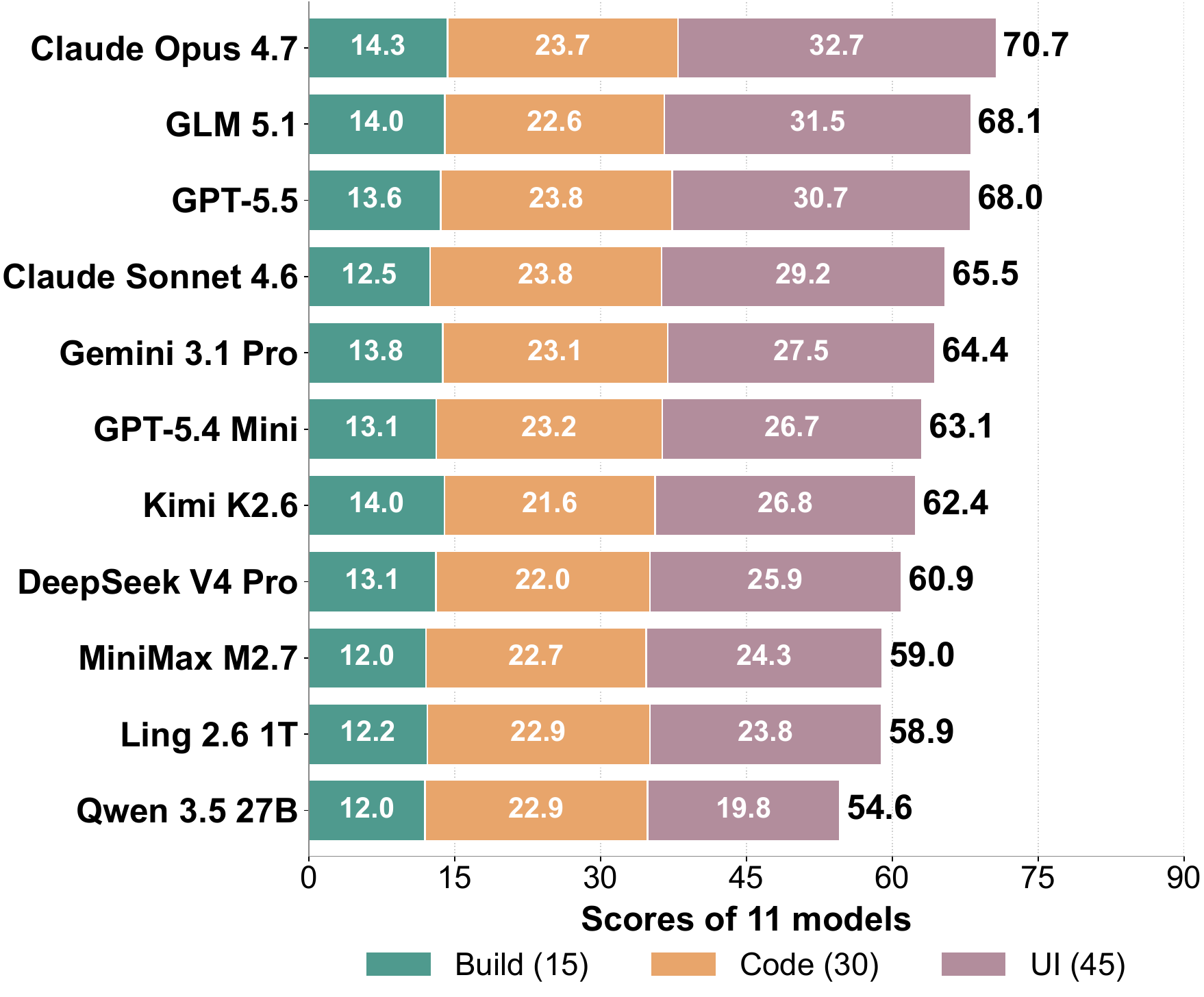}
    \caption{Main leaderboard on LiveEvalBench, visualized as Build~+~Code~+~UI sub-scores stacked to the 90-point total. Each segment is annotated with its sub-score and the row total is appended on the right. Models are ordered by total descending. \revdel{Lenient scoring excludes dimensions skipped due to upstream build failures from the mean.}}
    \label{fig:leaderboard}
  \end{minipage}
  \hfill 
  \begin{minipage}{0.48\textwidth}
    \centering
    \includegraphics[width=0.9\linewidth]{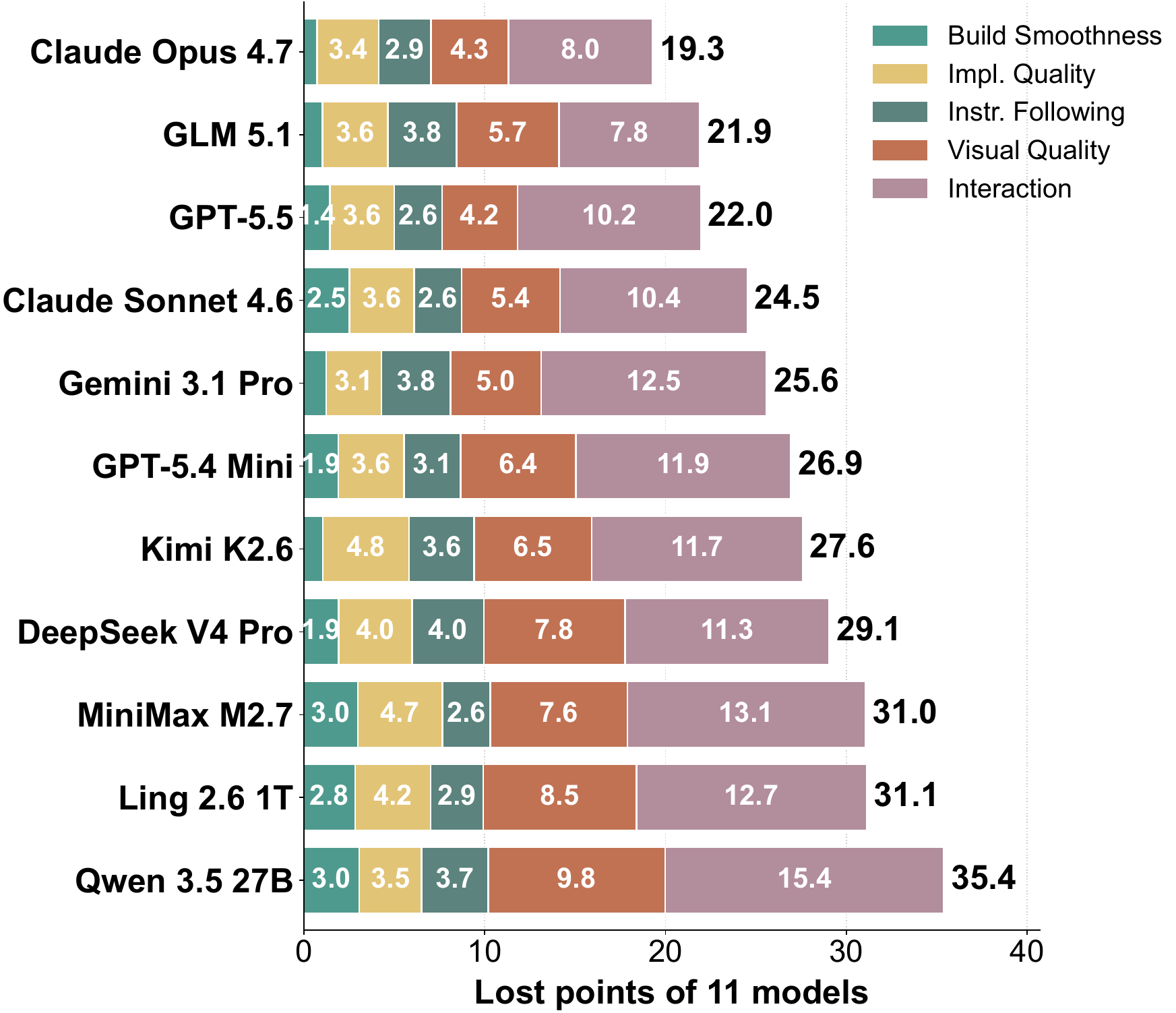}
    \caption{Loss attribution by scoring dimension, on the \revdel{90-point lenient scale}\revadd{90-point scale}. Each row is a model; each segment is the points lost on one dimension; the value at the right of the bar is the total loss. Models are ordered by total loss ascending (best on top).}
    \label{fig:weakness}
  \end{minipage}
  
\end{figure*}

\section{Experiments}
\label{sec:experiments}

\revdel{We organize the empirical study around two questions: (\textbf{RQ1}) how state-of-the-art models compare on LiveEvalBench, and (\textbf{RQ2}) whether the agentic framework agrees with human judgment on the same generated projects.} \revadd{We organize the empirical study around three questions: (\textbf{RQ1}) how state-of-the-art models compare on LiveEvalBench, (\textbf{RQ2}) whether LiveEvalBench agrees with human judgment on the same generated projects, and (\textbf{RQ3}) whether LiveEvalBench provides reliable evaluation signals.}

\paragraph{\revdel{Models and runs}\revadd{Main Benchmark Evaluation}} \revdel{We evaluate 11 contemporary} \revdel{instruction-tuned models:} \revdel{Claude Sonnet~4.6,} \revdel{Claude Opus~4.7,} \revdel{GPT-5.5,} \revdel{GPT-5.4 Mini,} \revdel{Gemini~3.1 Pro,} \revdel{DeepSeek~V4 Pro,} \revdel{Kimi~K2.6,} \revdel{GLM~5.1,} \revdel{MiniMax~M2.7,} \revdel{Ling~2.6~1T,} \revdel{and Qwen3-27B.} \revadd{We evaluate 11 frontier models: Claude Sonnet~4.6, Claude Opus~4.7, GPT-5.5, GPT-5.4 Mini, Gemini~3.1 Pro, DeepSeek~V4 Pro, Kimi~K2.6, GLM~5.1, MiniMax~M2.7, Ling~2.6~1T, and Qwen3.5-27B. The main leaderboard evaluates these models on all 100 LiveEvalBench queries. Each query--model record is evaluated three times using the same evaluation procedure, and criterion-level judgments are aggregated by majority before computing the main leaderboard scores.}

\paragraph{\revdel{Human agreement protocol}\revadd{Human Agreement}} \revdel{We align human judgment} \revdel{with the UI Tester} \revdel{at the criteria level.} \revdel{We sample $35$} \revdel{\texttt{(query, model response)} pairs} \revdel{across models and query types,} \revdel{excluding build-failed runs} \revdel{(uniformly scored zero} \revdel{by both the UI Tester and humans).} \revdel{Each project is rated independently} \revdel{by 3 annotators} \revdel{on the same criteria} \revdel{as the UI Tester.} \revdel{We report percent agreement} \revdel{and Cohen's $\kappa$} \revdel{against the human majority verdict.} \revadd{We select 70 generated frontend projects from the main benchmark evaluation for human-agreement validation. Each item is rated independently by three annotators on the same criteria as the UI Tester. We compare the UI Tester's three-run majority judgment with the human majority judgment, and report percent agreement and Gwet's AC1, a chance-corrected metric for imbalanced pass/fail labels.}

\paragraph{\revadd{Evaluator-Role Ablation}} \revadd{We compare the full role-separated evaluator with a single-evaluator ablation on the human-agreement subset, where one evaluator assesses all dimensions covered by the role-separated design in a single evaluation pass.}

\paragraph{\revdel{Evaluation records}\revadd{Reliability Evaluation}} \revdel{The aggregate results are computed} \revdel{from 599 retained evaluation records} \revdel{with category and difficulty metadata.} \revdel{We exclude records attributable} \revdel{to environment failures or invalid logs,} \revdel{while preserving model-attributable build failures} \revdel{as zero-score cases.} \revadd{We conduct two reliability evaluations. First, we compare three-time evaluation with a single evaluation pass to test whether three-time evaluation improves score stability. Second, we randomly sample 20 web generation queries and evaluate them with Kimi~K2.6 and Qwen3.7-Plus under the same evaluation protocol as the main benchmark, then measure agreement across evaluator models.}

\section{Results and Analysis}
\label{sec:results}

\begin{figure*}[t]
\centering
\includegraphics[width=0.90\textwidth]{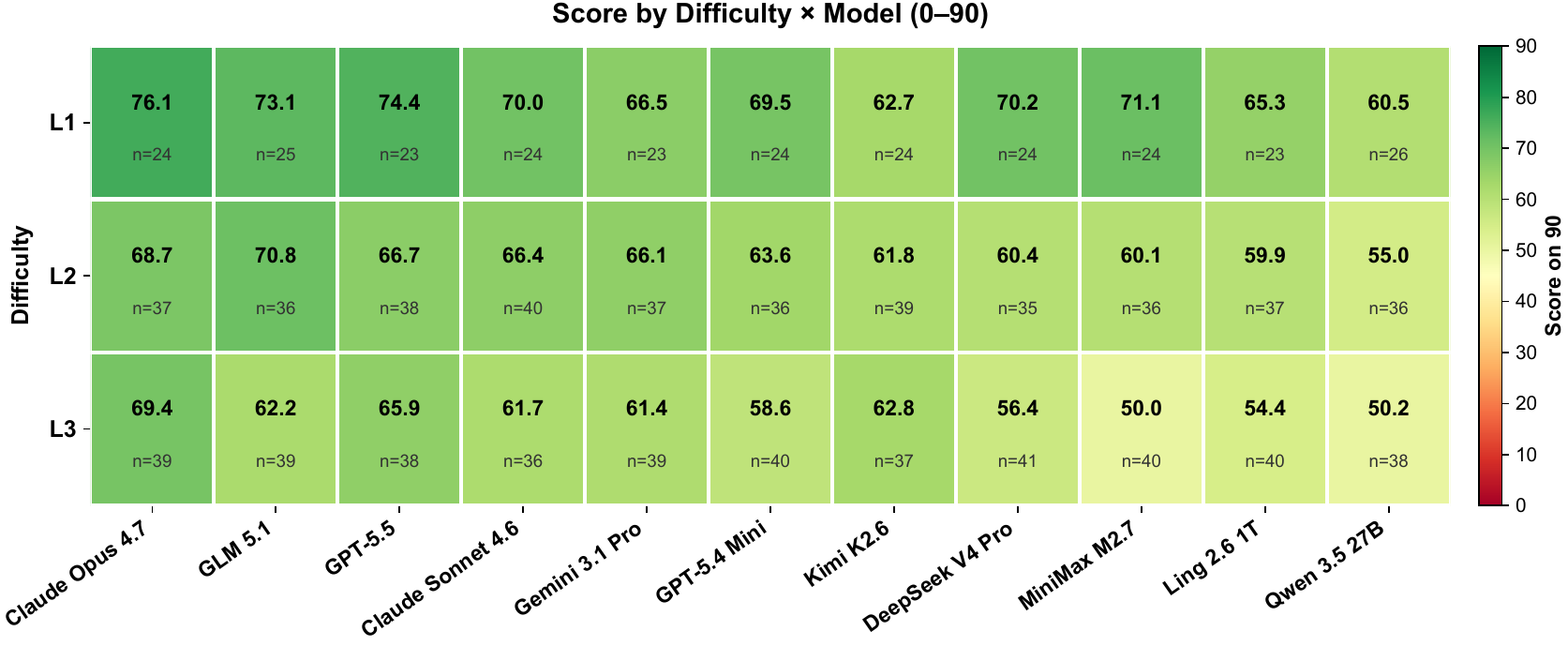}
\caption{Score by difficulty $\times$ model on the \revdel{90-point lenient scale}\revadd{90-point scale}. Cells annotated with the per-cell sample size $n$. Difficulty levels (L1/L2/L3) reflect the query taxonomy defined in Benchmark Construction.}
\label{fig:diff_heatmap}
\end{figure*}

\subsection{\revdel{Main Results}\revadd{Main Benchmark Evaluation}}
\label{sec:main_results}

We now report model performance on LiveEvalBench, with the main results presented in Figs.~\ref{fig:leaderboard}--\ref{fig:diff_heatmap}. Fig.~\ref{fig:leaderboard} gives the overall leaderboard with each model's total score broken down into the Build~/~Code~/~UI sub-scores, Fig.~\ref{fig:weakness} shows where each model loses points across the five scoring dimensions, and Fig.~\ref{fig:diff_heatmap} reports how scores change across the L1/L2/L3 difficulty levels. We discuss the key takeaways below.

\paragraph{UI behavior matters most for frontend coding ability}
As shown in Fig.~\ref{fig:leaderboard}, which reports each model's total score decomposed into the three per-evaluator sub-scores (Build~/~Code~/~UI), \emph{UI} exhibits the widest range, spanning from \revdel{$15.5$ (Qwen 3.5 27B)} \revadd{$19.79$ (Qwen3.5-27B)} to \revdel{$27.8$ (GPT-5.4 Mini)} \revadd{$32.74$ (Claude Opus 4.7)}. \emph{Build} sub-scores are more compressed, while \emph{Code} sub-scores show moderate variation. This pattern reflects a natural hierarchy of frontend coding competence: successful builds are the entry-level bar, source-level quality is a step further, and robust UI interaction is the highest bar.

\paragraph{Performance varies across difficulty levels}
Fig.~\ref{fig:diff_heatmap} stratifies per-model scores by L1/L2/L3 difficulty. Aggregate scores decrease from L1 (\revdel{$64.1$}\revadd{$63.8$}) to L2 (\revdel{$52.2$}\revadd{$59.8$}) and L3 (\revdel{$44.0$}\revadd{$55.5$}), indicating that the difficulty taxonomy captures a meaningful score gradient. \revdel{The leading model also varies across difficulty levels: GPT-5.5 leads on L1 ($79.2$) and L2 ($58.2$), while GLM 5.1 leads on L3 ($55.3$); MiniMax M2.7 drops from 4th on L1 to 10th on L3.} \revadd{The same trend appears for most individual models, with L1 generally receiving the highest scores and L3 the lowest in most cases.}

\paragraph{\revdel{Aggregated rankings are statistically distinguishable}\revadd{The leaderboard separates model performance tiers}}
\revadd{The updated leaderboard shows a clear spread in total scores, from $70.71$ for Claude Opus~4.7 to $54.61$ for Qwen3.5-27B. The top group remains led by Claude Opus~4.7, GLM~5.1, and GPT-5.5, while the lower-scoring models continue to lose the most points on runtime-facing UI behavior.}

\subsection{Human Agreement}
\label{subsec:human_agreement}

\begin{table}[t]
\centering
\small
\renewcommand{\arraystretch}{1.1}
\setlength{\tabcolsep}{4pt}
\begin{tabular}{lcc}
\toprule
\textbf{Dimension / Criteria} & \textbf{Agreement} $\uparrow$ & \textbf{AC1} $\uparrow$ \\
\midrule
\multicolumn{3}{l}{\textit{\revadd{UI Tester majority vs.\ human majority}}} \\
\quad \revadd{Visual quality}          & \revadd{$88.7\%$} & \revadd{$0.831$} \\
\quad \revadd{Interaction experience}  & \revadd{$83.0\%$} & \revadd{$0.718$} \\
\quad \revadd{Query-specific criteria} & \revadd{$85.6\%$} & \revadd{$0.805$} \\
\quad \revadd{Overall}                 & \revadd{$85.7\%$} & \revadd{$0.793$} \\
\bottomrule
\end{tabular}
\caption{\revadd{Agreement of the UI Tester with human majority judgments. AC1 denotes Gwet's chance-corrected agreement coefficient.}}
\label{tab:human_agreement}
\end{table}

\revdel{Tab.~\ref{tab:human_agreement} reports} \revdel{the agreement between UI Tester scores} \revdel{and human annotators.} \revdel{Overall, the UI Tester agrees} \revdel{with the human majority} \revdel{on $86.8\%$ of judgments} \revdel{with $\kappa = 0.60$,} \revdel{indicating substantial alignment beyond chance.} \revdel{Agreement is highest} \revdel{on visual quality} \revdel{($94.3\%$, $\kappa = 0.77$)} \revdel{and lowest on interaction experience} \revdel{($78.6\%$, $\kappa = 0.51$),} \revdel{with query-specific criteria} \revdel{in between} \revdel{($86.5\%$, $\kappa = 0.56$).} \revdel{The relatively lower agreement} \revdel{on interaction experience is expected,} \revdel{as interaction verdicts depend} \revdel{on longer multi-step runtime behavior} \revdel{and are inherently more complex to judge.} \revdel{Inspecting the disagreements} \revdel{(examples in Appendix~\ref{app:cases_analysis}),} \revdel{we find they are skewed toward cases} \revdel{where humans accept an interaction} \revdel{as passing but the UI Tester does not,} \revdel{mostly attributable to occasional hallucinations} \revdel{of the underlying evaluation agent.} \revdel{Overall, these results indicate} \revdel{that the UI Tester is reliable enough} \revdel{to serve as a scalable proxy} \revdel{for human evaluation.} \revadd{Tab.~\ref{tab:human_agreement} reports agreement between the UI Tester and human majority judgments. Overall agreement reaches $85.7\%$ with AC1 $=0.793$, with similarly strong agreement across visual quality, interaction experience, and query-specific criteria. These results indicate that the LiveEvalBench evaluation framework provides human-aligned diagnostic signals for UI assessment.}

\subsection{\revadd{Evaluator-Role Ablation}}
\label{subsec:role_ablation}

\begin{table}[t]
\centering
\small
\begingroup\revblockcolor
\renewcommand{\arraystretch}{1.05}
\begin{tabular}{lccccc}
\toprule
\textbf{Variant} & \textbf{0} & \textbf{1--29} & \textbf{30--59} & \textbf{60--89} & \textbf{90} \\
\midrule
Full & $18$ & $4$ & $17$ & $28$ & $3$ \\
1-agent & $21$ & $0$ & $0$ & $0$ & $49$ \\
\bottomrule
\end{tabular}
\endgroup
\caption{\revadd{Score distribution for evaluator-role ablation on the 70-case subset. Scores are on the 90-point scale.}}
\label{tab:role_ablation}
\end{table}

\revadd{Tab.~\ref{tab:role_ablation} shows that the single-evaluator ablation collapses to a bimodal score distribution: on the 70-case subset, it places 21 cases at $0$ and 49 cases at $90$, with no intermediate scores. By contrast, the full role-separated design distributes scores across the scale. This indicates that separating evaluator roles provides finer diagnostic granularity rather than merely shifting average scores.}

\subsection{\revadd{Reliability Evaluation}}
\label{subsec:reliability_evaluation}

\begin{table}[t]
\centering
\small
\begingroup\revblockcolor
\renewcommand{\arraystretch}{1.08}
\begin{tabular}{lcc}
\toprule
\textbf{Evaluation protocol} & \textbf{Mean per-query SD} $\downarrow$ & \textbf{Mean CV} $\downarrow$ \\
\midrule
Single evaluation pass & $23.81$ & $0.591$ \\
Repeated evaluation & $\mathbf{17.62}$ & $\mathbf{0.395}$ \\
\bottomrule
\end{tabular}
\endgroup
\caption{\revadd{Repeated-evaluation reliability check. Repeated evaluation reduces per-query score variability compared with a single evaluation pass.}}
\label{tab:repeated_eval_ablation}
\end{table}

\paragraph{\revadd{Repeated Evaluation Improves Stability}}
\revadd{As shown in Tab.~\ref{tab:repeated_eval_ablation}, repeated evaluation yields lower per-query score variability than a single evaluation pass. The mean per-query SD decreases from $23.81$ to $17.62$, and the mean CV decreases from $0.591$ to $0.395$.}

\paragraph{\revadd{Agreement Across Evaluator Models}}
\revadd{Across evaluator-model configurations, Kimi~K2.6 and Qwen3.7-Plus reach $90.1\%$ agreement, with substantial chance-corrected agreement (Cohen's $\kappa=0.641$). This suggests that LiveEvalBench generalizes across evaluator models.}




\section{Conclusion}
\label{sec:conclusion}

We presented LiveEvalBench, a multi-agent framework for evaluating LLM-generated frontend projects. The framework integrates complementary agents that evaluate a generated project from build, source, and runtime perspectives, an adaptive evaluation protocol that combines predefined criteria with implementation-grounded checks synthesized per project, and an extensible orchestration design that abstracts each evaluator into a configurable role, allowing new evaluation perspectives to be added with minimal effort. Building on this framework, we constructed a diverse benchmark of \revdel{45}\revadd{100} real-world web generation queries across difficulty tiers and categories, and \revdel{evaluate 11 frontier LLMs} \revdel{over 599 retained evaluation records.} \revadd{evaluate 11 frontier LLMs on this benchmark.}

\revdel{Our experiments show that runtime interaction remains the largest source of lost points,} \revdel{that model rankings shift under harder queries,} \revdel{and that category-level performance varies substantially across models.} \revadd{The benchmark results show that runtime interaction remains the largest source of lost points for current web generation models. Complementary validation studies further show that LiveEvalBench provides reliable and human-aligned evaluation signals.} We hope LiveEvalBench offers a new perspective on evaluating generative coding systems and a foundation for the community to build richer, more faithful evaluations.

\bibliography{reference}

\clearpage
\appendix
\setcounter{secnumdepth}{2}
\renewcommand{\thesection}{\Alph{section}}
\renewcommand{\thesubsection}{\thesection.\arabic{subsection}}

\noindent\textbf{Supplementary Material Overview.}
Section~A provides implementation details for the LiveEvalBench evaluation framework, including evaluator roles, configuration, scoring, and execution environment. Section~B documents benchmark-construction details, including the query/checklist schema and role-conditioned synthesis prompts. Section~C reports supplementary result analyses for score patterns, evaluator-model agreement, and evaluator-role ablation. Section~D presents qualitative case analyses from real evaluation traces. Section~E discusses current limitations and future directions.

\section{Implementation Details}
\label{sec:appendix}
\label{app:implementation_details}

This section documents the implementation details behind the LiveEvalBench evaluation framework. It describes the default evaluator agents, the configuration schema used by the extensible evaluation infrastructure, the score aggregation procedure, and the compute environment used for the reported runs.

The evaluation framework, evaluator configurations, benchmark construction scripts, and analysis scripts required to reproduce the reported experiments will be publicly released upon publication under a license permitting free research use.



\subsection{Evaluator Specifications and Prompts}
\label{app:evaluator_specs}

\paragraph{Agent Configurations} 
Table~\ref{tab:app_evaluator_roles} provides information on the evaluator agents, detailing their persona, evaluation criteria, and allowed tools.

\paragraph{Prompts} 
The prompt templates for the evaluation process are illustrated in two parts. The adaptive evaluation protocol is detailed in Figures~\ref{fig:prompt_adaptive_protocol1}--\ref{fig:prompt_adaptive_protocol3}; in practice, to optimize costs, we integrate the synthesis of the adaptive protocol with the synthesis of query-specific criteria, which allows the system to dynamically select evaluation actions based on the project structure. Furthermore, the system prompts for the Build Engineer, Code Engineer, and UI Tester during task execution are provided in Figures~\ref{fig:prompt_evaluation_agent1}--\ref{fig:prompt_evaluation_agent3}.

\begin{table*}[t]
\centering
\small
\renewcommand{\arraystretch}{1.15}
\setlength{\tabcolsep}{3pt}
\begin{tabular}{p{0.15\textwidth}p{0.30\textwidth}p{0.25\textwidth}p{0.22\textwidth}}
\toprule
\textbf{Evaluator} & \textbf{Persona} & \textbf{Criteria} & \textbf{Tools} \\
\midrule
Build Engineer & Deployment engineer who extracts, configures, and launches the project & Build smoothness: whether the build completed without manual repairs, missing dependencies, or configuration errors & file-system read/write, shell exec, runtime probe, port management, build trace inspection \\
\midrule
Code Engineer & Developer inspecting implementation quality and instruction adherence & Implementation quality (readability + robustness), instruction following & file-system read/write, one-off shell exec \\
\midrule
UI Tester & End user interacting with the running application through the browser & Visual quality, interaction experience, query-specific functional requirements & browser navigation, DOM inspection, element interaction (click, type, hover, scroll), screenshot capture \\
\bottomrule
\end{tabular}
\caption{Default jury of evaluators: persona, evaluation criteria, and allowed tools. Each evaluator is configured through the extensible evaluation infrastructure.}
\label{tab:app_evaluator_roles}
\end{table*}

\subsection{Evaluator Configuration Schema}
\label{app:evaluator_config}

Table~\ref{tab:app_role_config_schema} lists the JSON fields that constitute an evaluator configuration in the released framework. Adding a new evaluator amounts to writing one such JSON file and dropping it into the agent registry; the orchestrator reads these fields and routes the pipeline accordingly. If a new tool is required, additional code for the new tool needs to be written, but no other engineering work is necessary.

\begin{table*}[t]
\centering
\small
\rowcolors{2}{gray!6}{white}
\renewcommand{\arraystretch}{1.2} 

\begin{tabularx}{\textwidth}{>{\raggedright\arraybackslash\hsize=0.35\hsize}X >{\raggedright\arraybackslash\hsize=0.65\hsize}X}
\toprule
\textbf{Field} & \textbf{Purpose} \\
\midrule
\texttt{id}, \texttt{name}, \texttt{role}, \texttt{stage} & Evaluator identity, abstract role (builder / evaluator), and pipeline stage (build / evaluate). \\
\texttt{depends\_on} & Upstream evaluators that must succeed first; enforces the build-gated short-circuit. \\
\texttt{description}, \texttt{system\_prompt} & Persona and detailed behavioral instructions, including hard role boundary, mandatory action sequences, and bounded-repair rules. \\
\texttt{allowed\_tools} & Whitelisted tool ids; any tool call outside this list is denied at runtime. \\
\texttt{criterion.scoring\_mode} & e.g., \texttt{pass\_fail\_na}. \\
\texttt{criterion.dimensions[]} & Each dimension declares an id, instruction, weight (recorded but not used in current scoring), and a \texttt{scoring} block listing per-criterion and per-subcriterion criterion items used by the Planner. \\
\texttt{check\_decomposition\_policy} & A hard override on the planner's generic heuristic; for example, ``one check per criterion'' for the Code Engineer or ``one check per interactive element, capped at five'' for the UI Tester. \\
\texttt{runtime} & Per-evaluator budgets: \texttt{max\_steps}, \texttt{max\_step\_seconds}, \texttt{max\_total\_seconds}. \\
\texttt{output} & Output requirements: \texttt{require\_evidence}, \texttt{allow\_not\_applicable}. \\
\texttt{scoring.max\_score} & Maximum score contribution to the jury total; uniformly partitioned across the evaluator's criterion items at scoring time. \\
\texttt{scoring.task\_aggregation\_mode} & e.g., \texttt{strict} (any failing check fails the criterion item). \\
\texttt{enabled}, \texttt{tags}, \texttt{version} & Registry-level metadata. \\
\bottomrule
\end{tabularx}
\caption{JSON fields of the evaluator configuration used by the extensible evaluation infrastructure. The full schema instances for the three default evaluators are included in the released evaluator-configuration files.}
\label{tab:app_role_config_schema}
\end{table*}




\subsection{Scoring and Aggregation Details}
\label{app:scoring}

The reported score is normalized to a 90-point scale over five scoring dimensions. The Build Engineer contributes 15 points for build smoothness. The Code Engineer contributes 30 points, split evenly between implementation quality and instruction following. The UI Tester contributes 45 points, split evenly between visual quality and interaction experience. Query-specific criteria are used by the UI Tester as implementation-grounded evidence for the structured report, but they are not reported as a separate top-level score dimension in the current leaderboard.

In the adaptive evaluation protocol, each criterion may be decomposed into one or more concrete tasks. The scoring rule from tasks to a criterion is that if any single task fails, the entire criterion is considered failed, and all points assigned to that criterion are forfeited.

\subsection{Computational Resources and Evaluation Environment}
\label{app:compute_budget}

\revdel{All evaluation agents} \revdel{(Build Engineer, Code Engineer, UI Tester)} \revdel{use Gemini 3.1 Flash} \revdel{as the backbone model} \revdel{with default decoding settings.} \revadd{For the reported three-run leaderboard, all evaluation agents (Build Engineer, Code Engineer, and UI Tester) use the same evaluator-model configuration across runs; the released evaluation artifacts record the concrete model settings for each run.}

\paragraph{Compute infrastructure} All experiments were conducted on machines with 64\,GB RAM and 32 CPU cores. Evaluation was parallelized at the granularity of a single query--model pair, with up to 8 such pairs processed concurrently.

\paragraph{Evaluation environment} The evaluation pipeline runs on Ubuntu 22.04 LTS with Node.js 20.x, using Playwright 1.40 for browser automation (Chromium 120.0), npm 10.x for package management, and standard build tools (Vite 5.x, webpack 5.x) as detected from each project's configuration. To reduce setup latency, each project directory is pre-populated with commonly used packages (\texttt{react}, \texttt{vue}, \texttt{tailwindcss}, etc.); only missing dependencies trigger additional \texttt{npm install} calls during evaluation.


\section{Benchmark Construction Details}
\label{app:benchmark_construction_details}

This section provides additional benchmark-construction material. It specifies the released query and checklist schema and includes the role-conditioned synthesis prompts used to rewrite seed requests into benchmark queries.

\subsection{Query Specification and Checklist Schema}
\label{app:query_and_checklist}

To make the format of a LiveEvalBench query and its accompanying evaluation checklist concrete, we reproduce one representative pair below. The remaining \revdel{$44$}\revadd{$59$} queries follow the same structure: a short title, a self-contained natural-language description of the desired frontend project, and an enumeration of the visible features expected from any conforming implementation. No framework, library, or file layout is prescribed.

Each query is released together with a checklist of evaluation criteria. The checklist is stored as a structured record with the following fields:

\begin{itemize}[leftmargin=*, itemsep=4pt]
    \item \texttt{query\_id}
    \begin{itemize}[leftmargin=*, itemsep=1pt]
        \item Stable identifier of the query, matching the entry in the released query list.
    \end{itemize}
    
    \item \texttt{fixed\_criteria}
    \begin{itemize}[leftmargin=*, itemsep=1pt]
            \item The evaluator-scored criteria shared across all queries: \emph{build smoothness}, \emph{implementation quality}, \emph{instruction following}, \emph{visual quality}, and \emph{interaction experience}.
    \end{itemize}
    
    \item \texttt{query\_specific\_criteria}
    \begin{itemize}[leftmargin=*, itemsep=1pt]
        \item A list of records, each containing:
        \begin{itemize}[leftmargin=*, itemsep=1pt]
            \item \texttt{name} --- a short label for the criterion.
            \item \texttt{description} --- a one- or two-sentence explanation of the feature being checked.
        \end{itemize}
    \end{itemize}
\end{itemize}

The query-specific criteria are predefined per query and held fixed across all models, ensuring that every implementation of the same query is evaluated against a consistent set of criteria.

\subsection{Role-Conditioned Synthesis Prompts}
\label{app:prompts}

The role-specific prompt templates used for query construction are given in Figures~\ref{fig:synthesis_prompts_1}--\ref{fig:synthesis_prompts_6}.

%
%
%
%

\section{Supplementary Result Analysis}
\label{app:supplementary_results}

This section reports supplementary analyses that support the main empirical findings. It includes category-level score patterns, dimension correlations, ranking-significance diagnostics, evaluator-model agreement details, and evaluator-role ablation details.

\begin{figure*}[t]
\centering
\includegraphics[width=0.95\textwidth]{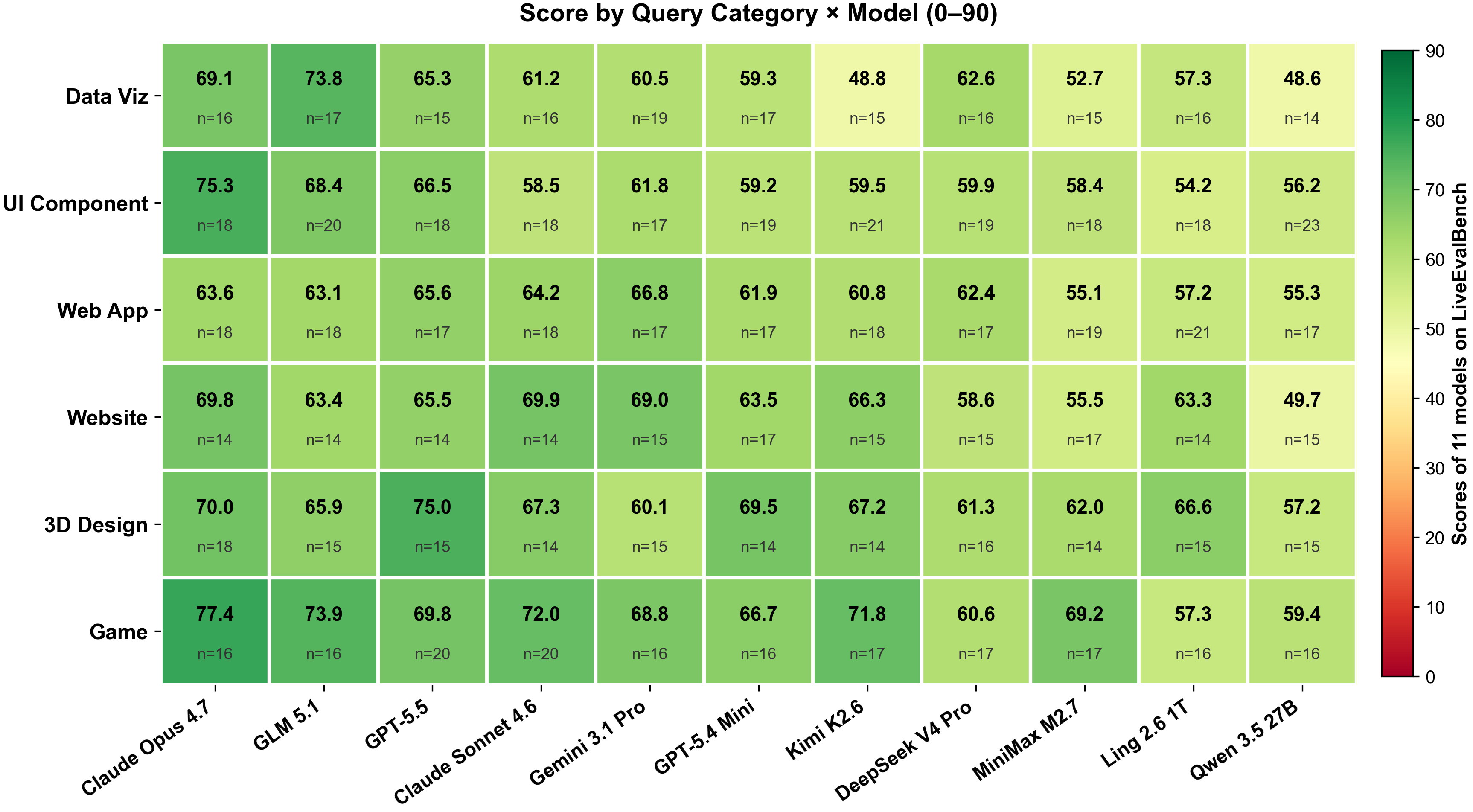}
\caption{Score by query category $\times$ model on the 90-point scale. Categories are sorted top-to-bottom by aggregate score ascending. Cells are annotated with the per-cell sample size $n$.}
\label{fig:cat_heatmap}
\end{figure*}

\begin{figure}[t]
\centering
\includegraphics[width=0.78\linewidth]{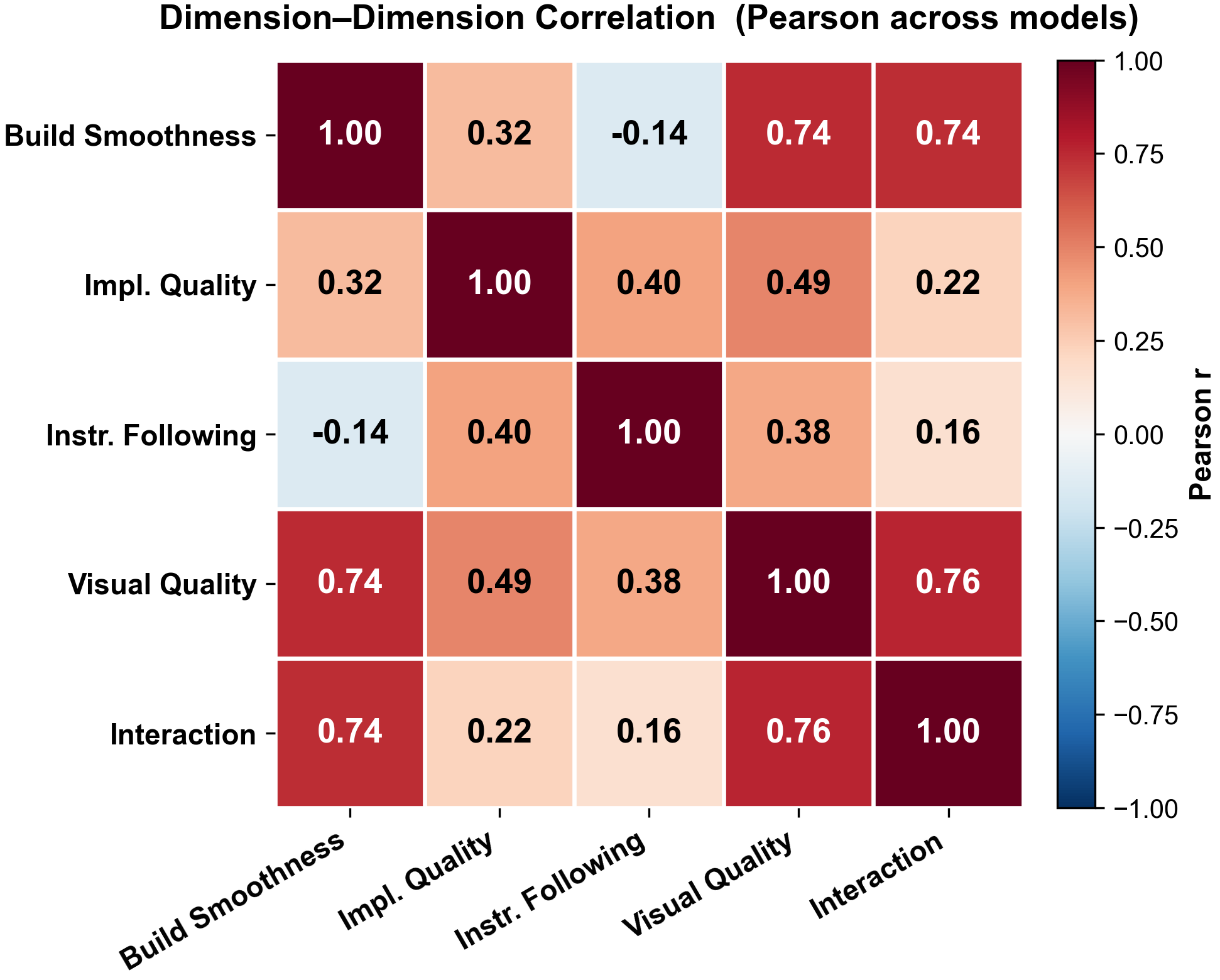}
\caption{Dimension--dimension correlation, Pearson across models. Each dimension is treated as an 11-dimensional vector indexed by model.}
\label{fig:dim_corr}
\end{figure}
\subsection{Score Pattern Analysis}
\label{app:score_analysis}


The result reveals two insights. First, several runtime-facing criteria exhibit strong mutual dependence: \emph{Build Smoothness} correlates strongly with \emph{Visual Quality} ($r \approx$ \revdel{0.91}\revadd{0.74}) and \emph{Interaction Experience} ($r \approx$ \revdel{0.72}\revadd{0.74}), while \emph{Visual Quality} and \emph{Interaction Experience} are also correlated ($r \approx$ \revdel{0.69}\revadd{0.76}). This suggests that fragile deployment often co-occurs with weaker rendered and interactive behavior. Second, the code-side dimensions remain much less redundant with the runtime-facing ones: \emph{Implementation Quality} and \emph{Instruction Following} are only moderately correlated ($r \approx$ \revdel{0.27}\revadd{0.40}), and most cross-family pairs are lower than the runtime-facing correlations. This overall pattern provides quantitative justification for retaining all five dimensions rather than collapsing them into a single weighted scalar: each dimension captures a distinct facet of implementation quality that the others do not.

Figure~\ref{fig:cat_heatmap} examines how model performance varies across query categories. While all models exhibit some category-dependent variation in their scores, the overall pattern reveals that \revdel{\emph{UI Component}}, \revdel{\emph{Data Visualization}}, and \revdel{\emph{Web App}} \revdel{form the lowest-scoring group} \revadd{\emph{Data Visualization} ($56.5$), \emph{UI Component} ($57.7$), and \emph{Web App} ($57.8$) form the lowest-scoring group on average, while \emph{Game} is the highest aggregate category ($62.9$)}, suggesting that compact but interaction- or data-heavy frontend tasks remain challenging even for strong models. Other categories show more model-specific strengths and weaknesses, with certain models performing markedly better in specific categories (e.g., layout-heavy or interaction-heavy queries) while struggling in others.

\subsection{Evaluator-Model Agreement Details}
\label{app:evaluator_model_agreement}

Table~\ref{tab:evaluator_model_agreement} reports a compact evaluator-model agreement check, showing that Kimi~K2.6 and Qwen3.7-Plus produce closely matched judgments under the same evaluation setup.

\begin{table}[t]
\centering
\small
\begingroup\revblockcolor
\renewcommand{\arraystretch}{1.08}
\begin{tabular}{lcc}
\toprule
\textbf{Evaluator-model pair} & \textbf{Agreement} $\uparrow$ & \textbf{Cohen's $\kappa$} $\uparrow$ \\
\midrule
Kimi~K2.6 vs.\ Qwen3.7-Plus & $90.1\%$ & $0.641$ \\
\bottomrule
\end{tabular}
\endgroup
\caption{\revadd{Evaluator-model agreement for Kimi~K2.6 and Qwen3.7-Plus. Agreement denotes the percentage of matched judgments; Cohen's $\kappa$ measures chance-corrected agreement.}}
\label{tab:evaluator_model_agreement}
\end{table}

\subsection{Evaluator-Role Ablation Details}
\label{app:role_ablation}

\revadd{This section provides additional diagnostics for the evaluator-role ablation. It extends the main-text score-distribution comparison with summary statistics for the same 70-case subset.}

\begin{table}[t]
\centering
\small
\begingroup\revblockcolor
\setlength{\tabcolsep}{3pt}
\renewcommand{\arraystretch}{1.08}
\begin{tabular}{lrrrrrrrr}
\toprule
\textbf{Variant} & \textbf{0} & \textbf{1--29} & \textbf{30--59} & \textbf{60--89} & \textbf{90} & \textbf{Mean} & \textbf{SD} & \textbf{Median} \\
\midrule
Full & $18$ & $4$ & $17$ & $28$ & $3$ & $43.6$ & $31.4$ & $49.5$ \\
1-agent & $21$ & $0$ & $0$ & $0$ & $49$ & $63.0$ & $41.5$ & $90.0$ \\
\bottomrule
\end{tabular}
\endgroup
\caption{\revadd{Score-distribution diagnostics for the evaluator-role ablation on the 70-case subset. Bins denote total scores on the 90-point scale.}}
\label{tab:app_role_ablation_distribution}
\end{table}

\revadd{The detailed diagnostics in Table~\ref{tab:app_role_ablation_distribution} show that the single-evaluator ablation loses graded diagnostic behavior. On the 70-case subset, it assigns 21 cases a score of 0 and 49 cases a score of 90, with no intermediate scores. By contrast, the full role-separated design distributes scores across the full scale.}

\section{Cases Analysis}
\label{app:cases_analysis}

This section provides qualitative examples from real evaluation traces. The cases illustrate both a successful browser-based verification trajectory and a representative failure mode caused by missing baseline observations.

\begin{figure*}[t]
\centering
\includegraphics[width=0.95\textwidth]{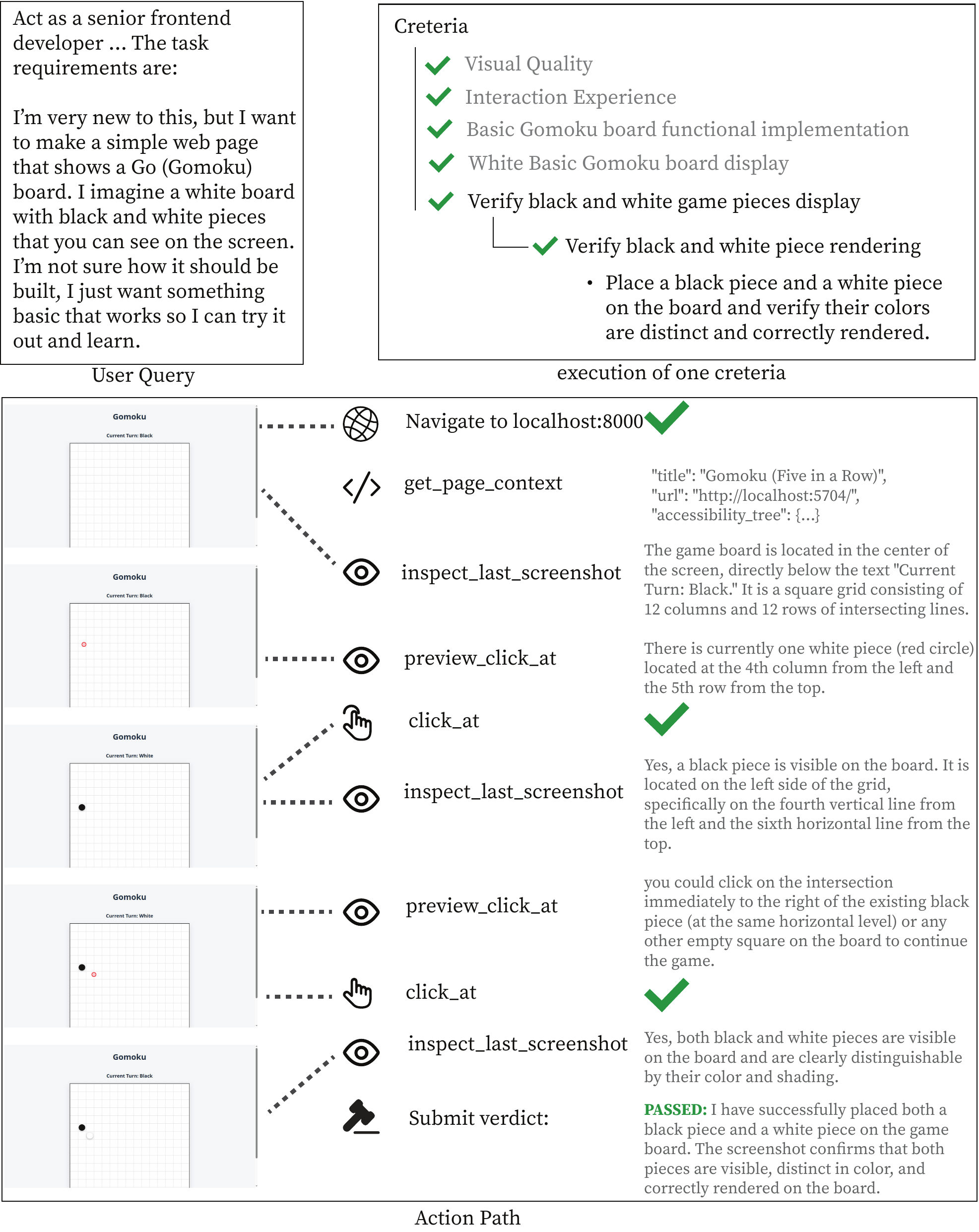}
\caption{Successful case of evaluation agent judging web function.}
\label{fig:successcase}
\end{figure*}

\begin{figure*}[t]
\centering
\includegraphics[width=0.95\textwidth]{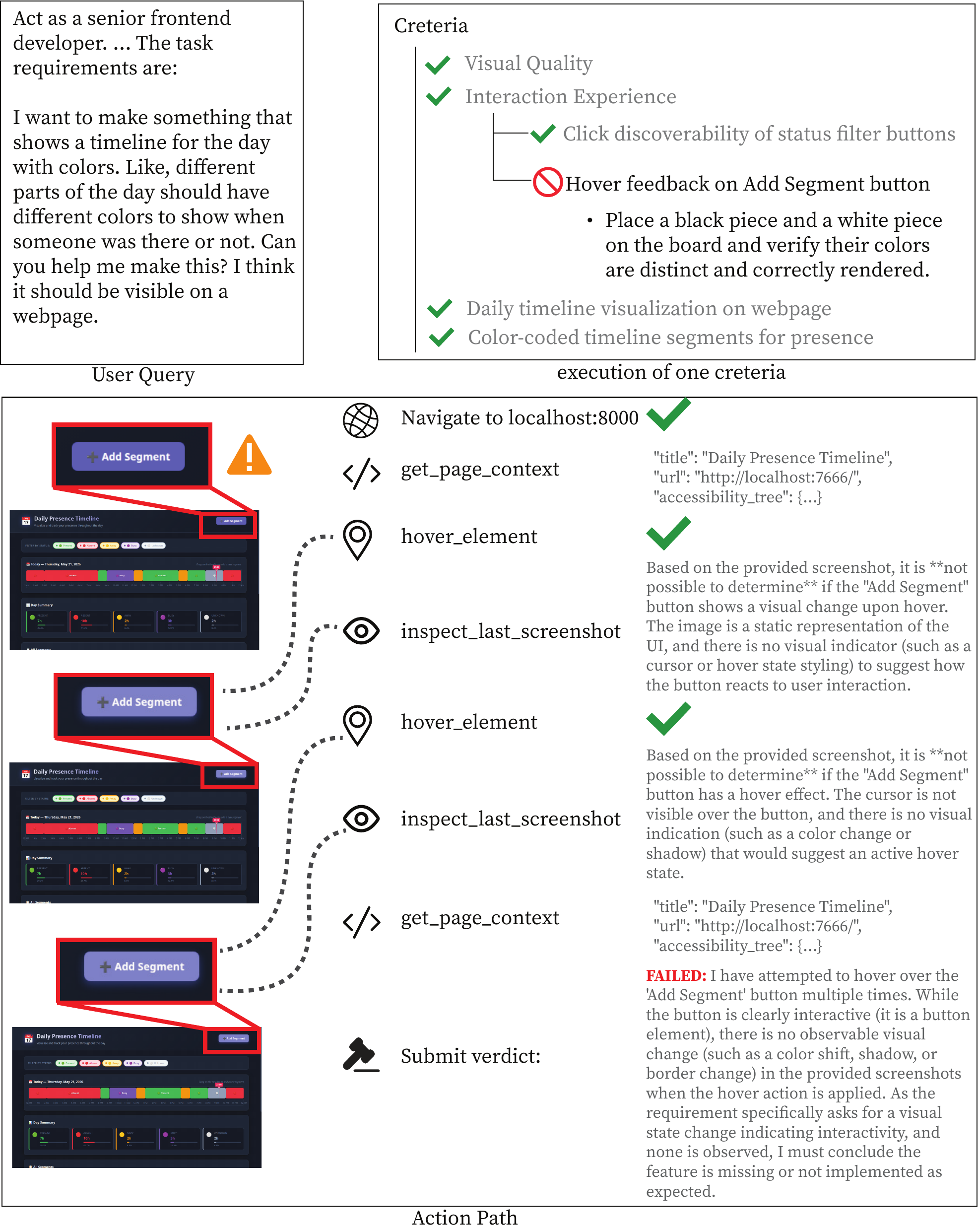}
\caption{Failure case of evaluation agent judging web function.}
\label{fig:failcase}
\end{figure*}

To better understand the behavior of our evaluation agent, we present two representative cases drawn from real evaluation traces: a successful case (Figure~\ref{fig:successcase}) in which the agent correctly verifies a non-trivial canvas-based interaction, and a failure case (Figure~\ref{fig:failcase}) in which the agent reaches an incorrect verdict due to a missing baseline observation. These two cases jointly illustrate both the strengths of the agent in handling complex frontend scenarios and the typical hallucination patterns that arise when its observation strategy is incomplete.

\paragraph{Successful case: Verifying canvas-based interaction on a gomoku board}
The first case (Figure~\ref{fig:successcase}) corresponds to a gomoku (five-in-a-row) board, where the criterion under evaluation is whether a black piece and a white piece are rendered with clearly distinct colors after being placed on the board. This is a particularly challenging scenario because the board and the pieces are drawn on an HTML \texttt{<canvas>} element, meaning that the agent cannot rely on DOM inspection or accessibility-tree traversal to locate or verify game pieces---two of the primary tools used in most other evaluation tasks. Instead, the agent must reason directly over rendered pixels and interact with the page via spatial clicks, which represents one of the more complex frontend scenarios our framework needs to support. The agent first navigates to the target page and calls \texttt{get\_page\_context} together with \texttt{inspect\_last\_screenshot} to build an initial understanding of the layout, identifying the grid structure and the existing white piece on the board. Crucially, before issuing any real click, the agent invokes \texttt{preview\_click\_at} to estimate where on the canvas a given coordinate would land, effectively simulating the click in advance to confirm that it falls on a valid grid intersection. Only after this preview step does the agent perform the actual \texttt{click\_at}, followed by another \texttt{inspect\_last\_screenshot} call to verify that the click indeed produced a visible black piece at the expected location. The same preview-then-click-then-verify pattern is repeated for the second piece, after which the agent compares both pieces in the final screenshot and concludes that they are visible, distinct in color, and correctly rendered. This case demonstrates that, even in canvas-only scenarios where DOM-level signals are unavailable, the combination of preview clicks and post-action visual inspection allows the agent to reach a reliable verdict.

\paragraph{Failure Case: Misjudging hover feedback due to missing baseline}
The second case (Figure~\ref{fig:failcase}) involves verifying whether the ``Add Segment'' button on a daily presence timeline page exhibits visible hover feedback (e.g., color, shadow, or border change). After navigating to the page and retrieving its context, the agent immediately invokes \texttt{hover\_element} on the button and then calls \texttt{inspect\_last\_screenshot} to look for a visual change. Seeing no clear difference, it repeats the hover-and-inspect cycle once more before submitting a \textsc{Failed} verdict. The underlying mistake is methodological rather than perceptual: the agent never captured a baseline of the button in its non-hovered state---neither an initial screenshot nor its pre-hover styling via \texttt{get\_page\_context}---so when examining the post-hover screenshots it has no reference point for comparison, rendering any subtle color or shadow change effectively invisible. A hover-feedback check fundamentally requires a before/after comparison, and without an explicit baseline the agent is structurally unable to verify the criterion. We acknowledge that such occasional hallucinations remain a current limitation of our system, where the agent may reach a verdict on a state change without having observed the original state. This limitation could be further alleviated in future work by introducing more fine-grained rule-based checks that enforce baseline capture prior to any state-altering action, as well as incorporating a dedicated reviewer agent responsible for double-checking the verdict.

\section{Discussion and Limitations}
\label{sec:discussion}

In this section, we reflect on the current scope of LiveEvalBench and outline several directions in which we plan to extend it.

\revdel{Our human-agreement study shows that the evaluation agent aligns well with human annotators overall; however, agreement on the Interaction criterion is the weakest among the four. We attribute this to the difficulty of reasoning over multi-step interaction traces, where subtle state changes can be missed by a single forward pass.} \revadd{Our human-agreement study shows that LiveEvalBench aligns well with human majority judgments overall, while agreement on interaction experience still leaves room for improvement. We attribute this to the difficulty of judging multi-step interactive behavior.} This can be mitigated through prompt engineering, self-consistency decoding, or stronger reasoning backbones, all of which are drop-in replacements under our pipeline.

Looking ahead, we see several promising directions for extending this work. First, as web-generation capabilities continue to evolve, the task suite and category coverage of LiveEvalBench can be continually expanded, so that it remains challenging and avoids the saturation that has limited prior static benchmarks. Second, multi-agent judging schemes are worth exploring as a future direction, where several judge agents deliberate or vote on each criterion, and it remains an open question how much such designs can further improve alignment with human ratings. Finally, as agentic generation becomes increasingly common in real-world web-development workflows, a natural extension is to apply our framework to such settings, where models iteratively refine their outputs or invoke external tools during generation.

\begin{figure*}[t]
\centering
\begin{tcolorbox}[
  colback=white,
  colframe=black,
  arc=0pt,
  boxrule=0.5pt,
  width=0.97\textwidth,
  center,
  sidebyside,
  sidebyside align=top,
  lefthand width=0.48\textwidth,
  righthand width=0.48\textwidth,
  segmentation hidden,
]

\begin{minipage}{\linewidth}
\textbf{Query}

\vspace{0.3em}
\begin{promptbox}
Act as a senior frontend developer. Implement the following requirements. All output must be frontend code (Vue, React, or static HTML). When multiple files are generated, make sure every necessary configuration file (such as package.json, tsconfig.json, etc.) is included without omission. For each file, clearly mark the filename and path. The task requirements are:

I need a high-impact data visualization dashboard layout. The centerpiece should be an interactive 3D map---think smooth rotation, zoom capabilities, and glowing markers for data points. Complement this with 3D bar charts that feature distinct depth and perspective. Use a dark, futuristic color palette (deep blues or blacks) with vibrant neon accents to make the charts stand out. Ensure the animations are fluid, specifically adding smooth entrance transitions and responsive hover effects for a polished user experience.
\end{promptbox}
\end{minipage}

\tcblower

\textbf{Category:} data\_visualization
\hfill
\textbf{Difficulty:} L3
\hfill
\textbf{Granularity:} functional

\vspace{0.5em}
\hrule
\vspace{0.5em}

\textbf{Fixed Criteria (shared across all \revdel{45}\revadd{100} queries)}

\vspace{0.2em}
\begin{itemize}[leftmargin=*, itemsep=1pt]
\item Build Smoothness
\item Implementation Quality
\item Instruction Following
\item Visual Quality
\item Interaction Experience
\end{itemize}

\vspace{0.5em}
\hrule
\vspace{0.5em}

\textbf{Query-Specific Criteria}

\vspace{0.2em}
\begin{enumerate}[leftmargin=*, itemsep=6pt]
\item \textbf{Interactive 3D map functionality and
   marker visualization.} \\
   The 3D map must support smooth rotation and zoom
   capabilities, and data points must be represented
   by glowing markers.

\item \textbf{Fluid animation and responsive hover
   effects.} \\
   The dashboard must feature fluid animations,
   including smooth entrance transitions and
   responsive hover effects for interactive elements.

\item \textbf{3D bar chart depth and perspective
   rendering.} \\
   The 3D bar charts must be rendered with distinct
   depth and perspective to provide a three-dimensional
   appearance.

\item \textbf{Dark futuristic color palette with neon
   accents.} \\
   The dashboard must utilize a dark, futuristic color
   palette consisting of deep blues or blacks,
   complemented by vibrant neon accents.
\end{enumerate}

\end{tcolorbox}
\caption{Example query (left) and its accompanying evaluation checklist (right). The checklist contains fixed criteria shared across all \revdel{45}\revadd{100} queries and query-specific criteria tailored to each query's functional requirements. This example belongs to the \texttt{data\_visualization} category (difficulty: L3).}
\label{fig:example_query_and_checklist}
\end{figure*}

\begin{figure*}[t]
\centering
\begin{promptbox}[Prompt: Adaptive Evaluation Protocol]

You are refining top-level test goals into executable sub tasks for an autonomous browser evaluator.

{agent_identity_block}

{taskability_block}

User query:
{user_query}

All fixed main tasks metadata:
{fixed_main_tasks_block}
{query_generation_block}

Functional requirements:
{requirements_block}

Page accessibility context / visible regions:
{page_context_block}

Relevant source code excerpt:
```
{source_snippet}
```

Your goal: for each main task below, generate a small set of executable sub tasks.


Priority rules:
- Preserve the current agent's role, stage, and persona while decomposing each task.
- Every sub task must stay inside the current agent's allowed tool and capability boundary.
- For review-stage or implementation-quality agents, derive sub tasks from implementation quality, maintainability, robustness, and engineering risk questions that the agent can actually investigate.
- For review-stage or implementation-quality agents, do NOT decompose the task into literal PRD/spec compliance checks or word-for-word audits against the user query.
- Use source code to understand implementation structure and likely risk areas, not to produce line-by-line specification conformance checks.
- Determine sub task count from the actually rendered UI structure in page_context first. Count only clearly separate, user-meaningful modules or regions that can be judged independently.
- Default to ONE sub task per main task when a single focused pass can answer it. Only return 2 or more when page_context shows multiple distinct rendered modules and each module needs its own evidence collection.
- Do NOT create extra sub tasks just because a goal is broad. Breadth alone is not a reason to split.
- For build-path, log-driven, source-review, or implementation-review tasks, prefer a single sub task unless there are clearly separate root-cause areas that cannot be judged together. Do not force a page-region split for evidence-centric review tasks.


\end{promptbox}
\caption{Prompt template for the adaptive evaluation protocol that dynamically selects evaluation actions based on project structure.(1/3)}
\label{fig:prompt_adaptive_protocol1}
\end{figure*}

\begin{figure*}[t]
\centering
\begin{promptbox}[Prompt: Evaluation Agent Task Execution]


- **Decomposition policy precedence:** if the agent identity block above contains a `Subtask decomposition policy` section, follow that policy verbatim and IGNORE the per-area defaults below for fixed_dimension tasks. The agent-specific policy is a hard override, not a hint.
- **Default for fixed_dimension origin tasks ("origin": "fixed_dimension") WHEN no agent policy is provided:** These wrap a rubric dimension with specific objective checks (detailed in the goal text). Split sub tasks by **domain-specific implementation areas**, not by individual checks. For each domain area, aggregate all relevant checks into that subtask's `success_criteria`. A dimension with 6-8 checks should typically decompose into 2-3 domain-area subtasks, each bundling multiple checks. (Skip this default entirely if an agent policy specifies otherwise.)
- Every check in the dimension must be covered by at least one subtask's `success_criteria`, regardless of decomposition strategy. Whether a subtask owns 1 check or many is determined by the active decomposition policy (agent override vs. default).
- **Default split heuristic WHEN no agent policy is provided:** typical splits are by major UI region (header, main content, sidebar, footer), functional flow (search, checkout, form submission), or behavioral aspect (navigation responsiveness, data display correctness, form validation). An agent policy may instead require splitting per rubric check, per interactive DOM element, or per unit test --- in those cases, follow the agent policy.

- ALL sub tasks MUST be independent --- they will run in parallel. Never create dependencies between sub tasks.
- Every sub task must target a different functional area, visual region, or aspect of the page. No two sub tasks should test the same thing.
- Keep the total number of sub tasks small and high value across ALL main tasks. Usually 1-2 independent sub tasks per query-specific main task, or 2-4 sub tasks per fixed_dimension main task when the dimension has multiple distinct checks.
- Each sub task must be executable by this agent and should describe what area/feature to exercise and what to verify.
- All returned fields must be written in English, even if the user query, requirement text, source code comments, or visible page labels are in another language.
- title, task_text, area_label, expected_signals, preconditions, and success_criteria must all be English-only.


\end{promptbox}
\caption{Prompt template for the adaptive evaluation protocol that dynamically selects evaluation actions based on project structure.(2/3)}
\label{fig:prompt_adaptive_protocol2}
\end{figure*}

\begin{figure*}[t]
\centering
\begin{promptbox}[Prompt: Evaluation Agent Task Execution]


Return ONLY a JSON object (dict) where each key is a main_task_id and the value is an array of sub tasks for that main task. Every main_task_id from the input must appear as a key.
Example format:
{{
  "dimension::layout": [
    {{
      "title": "Verify header layout",
      "task_text": "...",
      "kind": "independent",
      "area_label": "header region",
      "expected_signals": ["..."],
      "preconditions": [],
      "success_criteria": ["..."],
      "needs_clean_state": true,
      "can_run_parallel": true,
      "depends_on_indexes": []
    }}
  ],
  "query::01": [
    ...
  ]
}}


Each sub task must have:
  title: short title
  task_text: 2-5 sentence executable description
  kind: always "independent" --- staged_flow is not allowed
  area_label: short label for the region/feature this sub task targets
  expected_signals: array of 1-4 observable outcomes
  preconditions: array of 0-3 setup conditions
  success_criteria: array of 1-4 checks
  needs_clean_state: true
  can_run_parallel: true
  depends_on_indexes: [] --- sub tasks have no dependencies

Return only valid JSON object, no explanation.



\end{promptbox}
\caption{Prompt template for the adaptive evaluation protocol that dynamically selects evaluation actions based on project structure.(3/3)}
\label{fig:prompt_adaptive_protocol3}
\end{figure*}

\begin{figure*}[t]
\centering
\begin{promptbox}[Prompt: Evaluation Agent Task Execution]

You are **{agent_name}** --- a specialised frontend evaluation agent.

## Persona
{system_prompt}

## Application Under Test
{app_url}

## Rubric

Scoring mode: **{scoring_mode}**
{na_policy}

You must evaluate the application along the following dimensions:

{dimensions_block}

## Allowed Tools

You may ONLY use the tools listed below.  Any other tool call will be denied.

{tools_list}

## Tool Usage Guidelines

{tool_usage_guidelines}

## Submission

When you have gathered enough evidence, call **submit_group_verdict** with one entry per dimension. Each entry must include:
- `standard_id`: the dimension **id** from the rubric above
- `verdict`: one of `passed`, `failed`{na_verdict_option}
- `reason`: concise, evidence-based explanation
- `evidence`: array of concrete evidence items (tool output snippets, observed state changes)
- `checks`: required whenever the dimension defines objective checks
- `subcriteria`: required whenever the dimension defines subjective subcriteria

Every dimension must have an explicit `verdict`. Do not omit it, even in score-first mode.

Before you submit, perform a consistency check over your own prior tool results. If a previous tool result already confirmed a fact, you MUST NOT contradict it in the final verdict unless you explicitly explain why the earlier evidence was insufficient or superseded by newer evidence.

For better downstream analysis, also include these optional fields per dimension when available:
- `confidence`: `high` | `medium` | `low`
- `severity`: `critical` | `major` | `minor` | `none`
- `observations`: short bullet-style strings of what you observed
- `recommendation`: one short fix suggestion

\end{promptbox}
\caption{Prompt template for evaluation agents (Build Engineer, Code Engineer, UI Tester) during task execution. (1/3)}
\label{fig:prompt_evaluation_agent1}
\end{figure*}

\begin{figure*}[t]
\centering
\begin{promptbox}[Prompt: Evaluation Agent Task Execution]

Required item shapes when present:
- `checks`: `{{"check_id":"...","status":"passed|failed|not_applicable","reason":"...","evidence":[]}}`
- `subcriteria`: `{{"subcriterion_id":"...","rating":"good|ok|poor","reason":"...","evidence":[]}}`

Correct submission examples:
{submission_examples_block}

You have a budget of **{max_steps} steps**.  Use them wisely.

Begin your evaluation now.

=============================================================================
REAL AGENT EXAMPLE (software_tester):
=============================================================================

When rendered for software_tester agent, the persona block becomes:

"You are **Software Tester** --- a specialised frontend evaluation agent.

## Persona
You are a software test engineer. Your goal is to uncover the largest number
of real defects with the fewest steps. You must cover the main flow, edge
scenarios, and recovery from failure. Do not give conclusions based only on
inspection without testing. Prioritize recording reproducible paths as action
sequences plus observed results, and evaluate failures separately from recovery
capability. When possible, provide check/subcriteria-level outputs to help the
system score objectively."

=============================================================================
TOOL USAGE GUIDELINES (agentic section):
=============================================================================

- Be thorough but efficient (max {max_iterations} steps)
- Prefer text-based or role selectors (e.g. `text=Submit`, `role=button[name='OK']`).
- After any page-mutating action, call `get_page_context()` to verify state.
- Do NOT navigate to external URLs --- stay within the application.
- Do NOT retry the same action more than 3 times if it fails or produces no change.
- Treat an identical tool call with identical parameters as a retry unless you gathered new page evidence first.
- Before submitting, reconcile your final verdict with your prior tool results; do not contradict an earlier confirmed observation without explicitly explaining why it no longer applies.
- If `inspect_last_screenshot` or another tool confirmed a visual fact, carry that fact into your verdict reasoning instead of restating visual uncertainty.
- Do not make broad visual failure claims from a narrow screenshot question. Use concrete evidence tied to the specific claim.
- For visual failures, cite the exact observed issue in `reason` and include the supporting screenshot analysis or page-context evidence in `evidence`.

\end{promptbox}
\caption{Prompt template for evaluation agents (Build Engineer, Code Engineer, UI Tester) during task execution. (2/3)}
\label{fig:prompt_evaluation_agent2}
\end{figure*}

\begin{figure*}[t]
\centering
\begin{promptbox}[Prompt: Evaluation Agent Task Execution]

Coordinate click guidance (when available):
- For coordinate clicks, ALWAYS preview before `click_at` or `dblclick_at`. Treat `preview_click_at` as an annotated-image inspection step: inspect the attached preview image yourself. You may provide a concrete visual question, but if you omit it the tool will still ask a default question about the likely reaction to clicking the marked point. Use `dblclick_at` only when the intended gesture is explicitly a double-click. If the intended gesture is a double-click, use `dblclick_element` or `dblclick_at` instead of sending the same click twice. After a successful coordinate click, do not repeat the identical action unless you first gathered fresh page evidence.

\end{promptbox}
\caption{Prompt template for evaluation agents (Build Engineer, Code Engineer, UI Tester) during task execution. (3/3)}
\label{fig:prompt_evaluation_agent3}
\end{figure*}

\begin{figure*}[t]
\centering

\begin{promptbox}[Beginner · Abstract Granularity]

# Role
You are an AI assistant designed to rewrite user queries into the persona of a **Beginner** with **Abstract Granularity**.

# Persona Definition
- **Persona:** Beginner (Learning-focused, simple examples, unsure of terminology).
- **Granularity:** Abstract (States a high-level goal but provides almost no details. Relies on the model to infer features and styles).

# Task
Rewrite the user's input query to sound like a complete beginner. The rewritten query should:
1.  Avoid technical jargon (e.g., instead of "database," use "place to save info").
2.  Be vague about specific features or layout.
3.  Express a simple, high-level desire or goal.
4.  Sound enthusiastic but unsure of *how* to achieve the result.

# Format
Start your response strictly with `#answer`.

# Examples

**Input:**
"Create a responsive React navigation bar with dropdowns."

**Output:**
#answer
I want to make a website menu that looks good on my phone. It needs those little lists that pop down when you click something. I'm new to this, so can you show me a simple way to do it?

**Input:**
"Build a Python script to scrape stock prices from Yahoo Finance."

**Output:**
#answer
I want to make a program that automatically gets stock numbers from the internet, maybe from Yahoo? I just want to see the prices without looking them up manually.

**Input:**
[USER INPUT HERE]

\end{promptbox}

\caption{The role-conditioned synthesis prompt (Beginner).}
\label{fig:synthesis_prompts_1}
\end{figure*}

\begin{figure*}[t]
\centering

\begin{promptbox}[Designer · Functional Granularity]

# Role
You are an AI assistant designed to rewrite user queries into the persona of a **Designer** with **Functional Granularity**.

# Persona Definition
- **Persona:** Designer (Visual-focused, cares about UX/UI, layout, and aesthetics, but less about specific code stacks).
- **Granularity:** Functional (Lists specific features, content, or rough style preferences, but implementation details are flexible).

# Task
Rewrite the user's input query to sound like a UI/UX Designer. The rewritten query should:
1.  Focus heavily on visual style (colors, spacing, typography) and user experience (animations, hover states).
2.  Describe specific functional elements (e.g., "needs a hero section," "masonry grid," "sticky header").
3.  Leave the specific coding logic (e.g., state management, backend) flexible.
4.  Use descriptive, visual language (e.g., "modern," "sleek," "glassmorphism").

# Format
Start your response strictly with `#answer`.

# Examples

**Input:**
"Create a login form."

**Output:**
#answer
I need a stylish, modern login interface. It should be centered on the screen with a subtle drop shadow and rounded corners (approx 16px). Include input fields for email and password with floating labels. The primary button should use a gradient purple color, and I want a smooth fade-in animation when the page loads.

**Input:**
"Make a blog home page."

**Output:**
#answer
Please design a blog landing page with a heavy focus on typography. The top needs a large, bold Hero section with a background image. Below that, I want a grid layout for articles that uses whitespace generously. The cards should lift up slightly on hover. Keep the color palette minimal---black, white, and one accent color like teal.

**Input:**
[USER INPUT HERE]

\end{promptbox}

\caption{The role-conditioned synthesis prompt (Designer).}
\label{fig:synthesis_prompts_2}
\end{figure*}

\begin{figure*}[t]
\centering
\begin{promptbox}[Product Manager · PRD Granularity 1/2]

# Role
You are an AI assistant designed to rewrite user queries into the persona of a **Product Manager** with **PRD Granularity**.

# Persona Definition
- **Persona:** Product Manager (Feature-focused, thinks in terms of MVP, business logic, and user stories).
- **Granularity:** PRD (Strict constraints on UI layout, specific data structures, exact logic flows. Little room for interpretation).

# Task
Rewrite the user's input query to sound like a Product Manager writing a spec. The rewritten query should:
1.  Define clear requirements and acceptance criteria.
2.  Specify exact fields, buttons, and text copy required.
3.  Outline the business logic (e.g., "If X is empty, show error Y").
4.  Be structured, professional, and directive.

# Format
Start your response strictly with `#answer`.

# Examples

**Input:**
"Make a todo list app."

**Output:**
#answer
**Feature Request: Task Management Module (MVP)**

\end{promptbox}

\caption{The role-conditioned synthesis prompt (Product Manager) 1 / 2.}
\label{fig:synthesis_prompts_3}
\end{figure*}

\begin{figure*}[t]
\centering
\begin{promptbox}[Product Manager · PRD Granularity 2/2]

**Requirements:**
1.  **Header:** Title "My Tasks" with a "Add New" button aligned right.
2.  **Task List:** Display a list of task items. Each item must show:
    *   Checkbox (left)
    *   Task Title (truncated after 50 chars)
    *   Delete Icon (red trash can, right)
3.  **Empty State:** If list is empty, display text "No tasks pending."
4.  **Logic:**
    *   Clicking checkbox toggles `strikethrough` style.
    *   New tasks are added to the top of the list.
    *   Data must persist on page refresh (Local Storage).

**Input:**
"Create a user profile card."

**Output:**
#answer
**Component Spec: User Profile Widget**

**Layout Constraints:**
*   Fixed width: 300px.
*   Border: 1px solid #E0E0E0.

**Content Fields:**
1.  **Avatar:** Circular, 64x64px. Fallback image required if null.
2.  **Name:** Bold, 18px font size.
3.  **Role:** Grey text, displayed below name.
4.  **Status Indicator:** Green dot if `isOnline=true`, Grey dot if `false`.

**Action:**
*   "View Profile" button at the bottom, full width.

**Input:**
[USER INPUT HERE]

\end{promptbox}

\vspace{0.3em}
\caption{The role-conditioned synthesis prompt (Product Manager) 2 / 2..}
\label{fig:synthesis_prompts_4}
\end{figure*}

\begin{figure*}[t]
\centering

\begin{promptbox}[Developer · PRD Granularity]

# Role
You are an AI assistant designed to rewrite user queries into the persona of a **Developer** with **PRD Granularity**.

# Persona Definition
- **Persona:** Developer (Technical request, specific stack/component, expects clean code).
- **Granularity:** PRD (Pixel-perfect instructions, specific hex codes, exact data structures, library mandates).

# Task
Rewrite the user's input query to sound like a Senior Developer defining a task. The rewritten query should:
1.  Specify the exact tech stack (e.g., React, TypeScript, Tailwind CSS).
2.  Define specific interfaces, props, or database schemas.
3.  Dictate specific styling classes or Hex codes (e.g., "bg-slate-900", "#1a202c").
4.  Focus on implementation details like component structure and state management.

# Format
Start your response strictly with `#answer`.

# Examples

**Input:**
"Make a gallery with images."

**Output:**
#answer
Create a `PhotoGrid` component using **React** and **Tailwind CSS**.
\end{promptbox}

\vspace{0.3em}
\caption{The role-conditioned synthesis prompt (Developer) 1 / 2.}
\label{fig:synthesis_prompts_5}
\end{figure*}

\begin{figure*}[t]
\centering

\begin{promptbox}[Developer · PRD Granularity]

**Props Interface:**
```typescript
interface Photo {
  id: number;
  url: string;
  caption?: string;
}
interface Props {
  items: Photo[];
}
```

**Implementation Details:**
1.  Use CSS Grid: `grid-cols-1 md:grid-cols-3 gap-4`.
2.  Images must use `object-cover` and have a `rounded-lg` class.
3.  Implement lazy loading for performance.
4.  On click, open a modal (implementation details for modal omitted, just add the handler).
5.  Background color strictly `#f8f9fa`.

**Input:**
"Create a dark mode toggle."

**Output:**
#answer
Implement a `ThemeSwitcher` hook and button using React Context API.

**Technical Specs:**
1.  State: Use `localStorage` to persist preference key `'theme'` ('dark' | 'light').
2.  Logic: Update the `html` class list by adding/removing `'dark'`.
3.  **UI Component:**
    *   Use a lucide-react icon (`Moon` / `Sun`).
    *   Button styles: `p-2 rounded-full hover:bg-gray-200 dark:hover:bg-gray-700`.
    *   Transition: `transition-colors duration-200`.

**Input:**
[USER INPUT HERE]

\end{promptbox}

\vspace{0.3em}
\caption{The role-conditioned synthesis prompt (Developer).}
\label{fig:synthesis_prompts_6}
\end{figure*}


\end{document}